\documentclass[acmsmall,nonacm]{acmart}
\AtBeginDocument{
  }

\usepackage{siunitx}
\usepackage{url}
\usepackage{enumitem}
\usepackage{algorithm}
\usepackage{algorithmic}
\usepackage{subcaption}
\usepackage{multirow}
\usepackage{pifont}
\usepackage{colortbl}
\usepackage[capitalise,nameinlink,noabbrev]{cleveref}
\usepackage{xspace}
\usepackage{listings}

\newcommand{\dataset}{LMOD+\xspace}

\begin{document}

\title{\dataset: A Comprehensive Multimodal Dataset and Benchmark for Developing and Evaluating Multimodal Large Language Models in Ophthalmology}

\author{Zhenyue Qin}
\authornote{Both authors contributed equally to this research}
\email{kf.zy.qin@gmail.com}
\affiliation{
  \institution{School of Medicine, Yale University}
  \city{New Haven}
  \state{CT}
  \postcode{06520}
  \country{USA}
}

\author{Yang Liu}
\authornotemark[1]
\email{yang.liu1082@gmail.com}
\affiliation{
  \institution{School of Computing, Australian National University}
  \city{Canberra}
  \state{ACT}
  \postcode{2601}
  \country{Australia}
}

\author{Yu Yin}
\email{yinyu201906@gmail.com}
\affiliation{
  \institution{School of Engineering, Imperial College London}
  \city{London}
  \postcode{SW7 2AZ}
  \country{UK}
}

\author{Jinyu Ding}
\email{jinyu.ding@yale.edu}
\affiliation{
  \institution{School of Medicine, Yale University}
  \city{New Haven}
  \state{CT}
  \postcode{06520}
  \country{USA}
}

\author{Haoran Zhang}
\email{casper.zhang@yale.edu}
\affiliation{
  \institution{School of Medicine, Yale University}
  \city{New Haven}
  \state{CT}
  \postcode{06520}
  \country{USA}
}

\author{Anran Li}
\email{anran.li@yale.edu}
\affiliation{
  \institution{School of Medicine, Yale University}
  \city{New Haven}
  \state{CT}
  \postcode{06520}
  \country{USA}
}

\author{Dylan Campbell}
\email{dylan.campbell@anu.edu.au}
\affiliation{
  \institution{School of Computing, Australian National University}
  \city{Canberra}
  \state{ACT}
  \postcode{2601}
  \country{Australia}
}

\author{Xuansheng Wu}
\email{xw54582@uga.edu}
\affiliation{
  \institution{School of Computing, University of Georgia}
  \city{Athens}
  \state{GA}
  \postcode{30602}
  \country{USA}
}

\author{Ke Zou}
\email{zou_ke@nus.edu.sg}
\affiliation{
  \institution{Yong Loo Lin School of Medicine, National University of Singapore}
  \postcode{119228}
  \country{Singapore}
}

\author{Tiarnan D. L. Keenan}
\email{tiarnan.keenan@nih.gov}
\affiliation{
  \institution{National Eye Institute, National Institutes of Health}
  \city{Bethesda}
  \state{MD}
  \postcode{20894}
  \country{USA}
}

\author{Emily Y. Chew}
\email{echew@nei.nih.gov}
\affiliation{
  \institution{National Eye Institute, National Institutes of Health}
  \city{Bethesda}
  \state{MD}
  \postcode{20894}
  \country{USA}
}

\author{Zhiyong Lu}
\email{luzh@ncbi.nlm.nih.gov}
\affiliation{
  \institution{National Library of Medicine,
National Institutes of Health}
  \city{Bethesda}
  \state{MD}
  \postcode{20894}
  \country{USA}
}

\author{Yih Chung Tham}
\email{thamyc@nus.edu.sg}
\affiliation{
  \institution{Yong Loo Lin School of Medicine, National University of Singapore}
  \postcode{119228}
  \country{Singapore}
}

\author{Ninghao Liu}
\email{ninghao.liu@uga.edu}
\affiliation{
  \institution{School of Computing, University of Georgia}
  \city{Athens}
  \state{GA}
  \postcode{30602}
  \country{USA}
}

\author{Xiuzhen Zhang}
\email{xiuzhen.zhang@rmit.edu.au}
\affiliation{
  \institution{School of Computing Technologies, RMIT University}
  \city{Melbourne}
  \state{VIC}
  \postcode{3000}
  \country{Australia}
}

\author{Qingyu Chen}
\authornote{Correspondence to: qingyu.chen@yale.edu}
\email{qingyu.chen@yale.edu}
\affiliation{
  \institution{School of Medicine, Yale University}
  \city{New Haven}
  \state{CT}
  \postcode{06520}
  \country{USA}
}

\renewcommand{\shortauthors}{Zhenyue Qin et al.}

\begin{abstract}
The rising prevalence of vision-threatening eye diseases poses a major global health and economic burden, yet timely diagnosis remains limited by workforce shortages, diagnostic delays, and restricted access to specialized care. Artificial intelligence (AI) offers potential solutions. In particular, recent progress in foundation models and large language models—especially multimodal large language models (MLLMs)—has shown promise in medical image interpretation and automated clinical documentation. However, advancing MLLMs for ophthalmology is hindered by the lack of unified, comprehensive benchmark datasets for development and evaluation. Most existing benchmarks were designed for earlier models, which focused on narrow tasks or specific disease conditions. These benchmarks typically provide outputs in the form of disease labels rather than free-text responses. As a result, they are less suitable for assessing emerging generative models.

In this work, we present \dataset, a large-scale multimodal ophthalmology benchmark dataset comprising 32,633 instances with multi-granular annotations across 12 common ophthalmic conditions and 5 imaging modalities. 
The dataset integrates imaging, anatomical structures, demographics, and free-text annotations. It supports primary ophthalmic applications such as anatomical structure recognition, disease screening, disease staging, and demographic prediction for potential performance bias evaluation. Alongside the dataset, we introduce a systematic and unified data curation pipeline that repurposes existing or new datasets for MLLM development.

\dataset extends our preliminary LMOD benchmark—the first multimodal ophthalmology benchmark for MLLMs—with three major enhancements. First, we expanded the dataset by nearly 50\% (from 21,933 to 32,633 instances). The color fundus photography (CFP) modality, the most accessible imaging modality in ophthalmology, was significantly enlarged to cover a broader range of pathological conditions. Second, we broadened task coverage to include (a) 12 binary disease diagnosis tasks for prevalent conditions such as diabetic retinopathy, age-related macular degeneration, and retinal vein occlusion; (b) multi-class ophthalmic disease diagnosis; (c) disease severity classification, including a diabetic retinopathy staging task, which uses two internationally adopted grading standards: the international clinical diabetic retinopathy classification and the Scottish diabetic retinopathy grading scheme classification; and (d) demographic prediction (age and sex) to assess potential model bias. Third, we systematically evaluated 24 state-of-the-art MLLMs, including recent models from the InternVL, Qwen, and DeepSeek families.

Our evaluations highlight both the promise and limitations of current MLLMs in ophthalmology. For example, Qwen-7B and InternVL achieved accuracies of 58.26\% and 57.83\% in disease screening under a zero-shot setting with a single model—a considerably more challenging paradigm than traditional fine-tuning, where separate models are trained for each specific task. InternVL also demonstrated potential in anatomical recognition. Nonetheless, overall performance remained suboptimal and often close to random baselines for challenging tasks such as disease staging, underscoring the substantial gap between general-domain MLLMs and the specialized requirements of ophthalmology.

We publicly release the dataset, curation pipeline, and leaderboard to encourage community-wide development and evaluation of MLLMs, with the goal of advancing ophthalmic applications and ultimately reducing the global burden of vision-threatening diseases through AI.
The dataset website, benchmark leaderboard, and download link are available at \href{https://kfzyqin.github.io/lmod_plus}{\url{https://kfzyqin.github.io/lmod_plus}}. \end{abstract}

\ccsdesc[500]{Applied computing~Health care information systems}
\ccsdesc[500]{Applied computing~Health informatics}
\ccsdesc[500]{Computing methodologies~Natural language processing}
\ccsdesc[500]{Computing methodologies~Computer vision}

\keywords{Multimodal large language models, ophthalmology, medical AI, benchmark dataset, healthcare computing}

\received{15 September 2025}

\maketitle

\section{Introduction}

The rising prevalence of vision-threatening eye diseases poses a major public health burden. In the United States alone, more than 90 million people are at high risk for vision loss~\cite{Saydah2020}, yet many remain undiagnosed or are diagnosed too late for effective treatment. For example, up to 50\% of patients with diabetic retinopathy do not receive timely eye examinations or are only identified at a stage when treatment is no longer effective~\cite{Chong503}. Surveillance studies report a median diagnostic delay of 22 weeks, with more than 70\% of affected patients experiencing permanent vision loss~\cite{Foot2017}.

Globally, vision impairment affects more than 2.2 billion people, with cataracts, age-related macular degeneration, glaucoma, and diabetic retinopathy accounting for nearly half of all cases. Yet, only 17–36\% of individuals with vision impairment receive appropriate interventions, highlighting a critical gap in timely screening and management~\cite{tham2014global,neely2017prevalence,cavan2017diabetic,2023_eye_WHO}. Key barriers include the time burden of manual examinations and documentation in ophthalmic clinics, as well as limited access to eye care in resource-constrained settings. The global economic impact is substantial, with preventable vision impairment contributing to an estimated \$411 billion in annual productivity loss~\cite{2023_eye_WHO}.

Artificial intelligence (AI) offers promising solutions to these challenges. Earlier approaches based on convolutional neural networks (CNNs), which automatically map medical image features to disease labels with supervised fine-tuning, have demonstrated strong performance in eye disease diagnosis~\cite{Ejaz2025}. More recently, pioneering studies on foundation models and large language models (LLMs)—particularly multimodal large language models (MLLMs)—have shown promise in medical image interpretation and automated clinical documentation. Compared to earlier models, MLLMs show robust zero-shot and few-shot learning capabilities. This allows them to perform effectively with minimal training samples and without the need for extensive task-specific fine-tuning. As a result, they are well-suited for resource-limited settings~\cite{liu2023summary, tian2024opportunities, de2023chatgpt}.

Despite their promise, a major challenge in advancing MLLMs for ophthalmology is the lack of unified, comprehensive benchmarks for development and evaluation. Most existing benchmarks were designed for earlier models such as CNNs, focusing on narrow tasks or specific disease conditions for fine-tuning. Moreover, these benchmarks typically provide outputs in the form of disease labels rather than free-text responses, making them less suitable for assessing the generative and reasoning capabilities of recent models.

More recent benchmarks tailored to newer models have primarily emphasized text-based tasks, such as general ophthalmology knowledge tests in multiple-choice format~\cite{2024_benchnephrology_nejmai, 2023_evalgptophth_ophsci}. While effective for evaluating purely language-based models, these benchmarks fail to reflect real-world ophthalmic practice, where medical imaging is indispensable. In practice, ophthalmic diagnosis requires integrating visual information from key imaging modalities such as fundus photography and optical coherence tomography alongside clinical history and examination findings~\cite{khan2021global}. Text-only benchmarks overlook the rich visual patterns that are critical for detecting the progression of diabetic retinopathy, changes in the glaucomatous optic disc, and features of macular degeneration.
Pioneering efforts to extend benchmarks to MLLMs in ophthalmology have begun to address these limitations. However, the scope of visual modalities remains narrow, often restricted to single data types such as surgical scenes~\cite{2024_cataract1k_scidata} or to single tasks such as region segmentation~\cite{2024_harvardfairseg_tmi}, limiting their ability to comprehensively assess model performance across diverse clinical scenarios.

This work extends our preliminary study,\footnote{\url{https://aclanthology.org/2025.findings-naacl.135/}}
 in which we introduced LMOD, the first large-scale multimodal ophthalmology benchmark dataset, and evaluated selected MLLMs on three tasks. Here, we present a significantly enhanced version, \dataset, comprising 32,633 images with multi-granular annotations across 12 common ophthalmic conditions and 5 imaging modalities. 
 The dataset encompasses color fundus photographs (CFP, 43.2\%), scanning laser ophthalmoscopy (SLO, 30.6\%), optical coherence tomography (OCT, 11.8\%), lens photographs (LP, 7.5\%), and surgical scenes (SS, 6.9\%). Patient demographics reveal a female predominance (60.6\%) versus male representation (39.4\%).
 This work introduces three key changes:

\begin{itemize}
\item We increased the dataset by nearly 50\%, from 21,933 to 32,633 images. In particular, we substantially enlarged the CFP modality—the most accessible imaging modality in ophthalmology—covering a broader range of pathological conditions detectable through CFP.
\item Beyond the original three tasks, we now include: (a) 12 binary eye condition diagnosis tasks covering prevalent diseases such as diabetic retinopathy, age-related macular degeneration, and retinal vein occlusion; (b) multi-class ophthalmic disease diagnosis; (c) disease severity classification, including both macular hole and diabetic retinopathy staging; and (d) demographic prediction (patient age and sex) to quantify potential bias in MLLMs.
\item We nearly doubled the number of evaluated MLLMs from 13 to 24, including recent state-of-the-art models such as the InternVL~\cite{2024_internvl_cvpr}, Qwen~\cite{bai2023qwen}, and DeepSeek series~\cite{wu2024deepseekvl2mixtureofexpertsvisionlanguagemodels}. To foster continued progress, we publicly release the updated full dataset, LMOD+ subset (a sampled 1000-instance representative subset) and introduce a dynamic leaderboard based on the subset to support ongoing benchmarking and model development in ophthalmology.
\end{itemize}

Using LMOD+, we systematically evaluated 24 state-of-the-art MLLMs. The results reveal heterogeneous performance across tasks: Qwen-7B and InternVL 2.5-8B showed potential in eye disease screening, achieving overall accuracies of 58.26\% and 57.83\% under the zero-shot setting with a single model, respectively, while InternVL 1.5-4B excelled in anatomical recognition tasks. Overall, our findings highlight a substantial gap between the performance of both general-domain and medical-domain MLLMs in ophthalmology and the specialized requirements of the field, underscoring the pressing need to develop and evaluate domain-specific MLLMs. To support further progress, we publicly release LMOD+ together with its curation and evaluation pipeline, which can be readily applied to emerging datasets and models. We encourage broader community efforts in the development and evaluation of MLLMs to advance ophthalmic applications and ultimately reduce the global burden of vision-threatening diseases with the assistance of AI.

\section{Related Work}
\begin{table*}[!t]
\centering
\caption{Comparison of existing general-domain and ophthalmology-specific benchmarks for evaluating large vision-language models, highlighting their supported modalities, coverage of image types, and evaluation perspectives.}
\setlength\tabcolsep{6pt}
\resizebox{1\linewidth}{!}{
\begin{tabular}{@{}lccccccccccc@{}}
\toprule
\multirow{3}{*}{\textbf{Benchmarks}} & \multicolumn{2}{c}{\textbf{Modalities}} & \multicolumn{5}{c}{\textbf{Image Types}} & \multicolumn{2}{c}{\textbf{Evaluation Perspectives}} \\
\cmidrule(lr){2-3} \cmidrule(lr){4-8} \cmidrule(l){9-10} 
& \multirow{2}{*}{Images} & \multirow{2}{*}{Texts} & Surgical Scenes & Optical Coherence & Scanning Laser & Lens Ophthalmoscopy & Color Fundus & Anatomical & Diagnosis \\
& & & (SS) & Tomography (OCT)
&  (SLO)
& Photographs (LP) & Photographs (CFP) & Understanding & Analysis \\
\midrule
\multicolumn{10}{c}{\textbf{General-Domain Benchmarks}} \\
\midrule
MMMU~\cite{2024_mmmu_cvpr} & \ding{51} & \ding{51} & \ding{55} & \ding{55} & \ding{55} & \ding{51} & \ding{51} & \ding{55} & \ding{55} \\
MME-RealWorld~\cite{2024_mme} & \ding{51} & \ding{51} & \ding{55} & \ding{55} & \ding{55} & \ding{55} & \ding{55} & \ding{55} & \ding{55} \\
UNK-VQA~\cite{2024_unkvqa_tpami} & \ding{51} & \ding{51} & \ding{55} & \ding{55} & \ding{55} & \ding{55} & \ding{55} & \ding{55} & \ding{55} \\
MMCBench~\cite{2024_benchlmmcor} & \ding{51} & \ding{51} & \ding{55} & \ding{55} & \ding{55} & \ding{55} & \ding{55} & \ding{55} & \ding{55} \\
MathVista~\cite{2023_mathvista_iclr} & \ding{51} & \ding{51} & \ding{55} & \ding{55} & \ding{55} & \ding{55} & \ding{55} & \ding{55} & \ding{55} \\
SEED-Bench~\cite{2024_seed_cvpr} & \ding{51} & \ding{51} & \ding{55} & \ding{55} & \ding{55} & \ding{55} & \ding{55} & \ding{55} & \ding{55} \\
\midrule
\multicolumn{10}{c}{\textbf{Ophthalmology-Specific Benchmarks}} \\
\midrule
Eval-GPT-Ophth~\cite{2023_evalgptophth_ophsci} & \ding{55} & \ding{51} & \ding{55} & \ding{55} & \ding{55} & \ding{55} & \ding{55} & \ding{55} & \ding{55} \\
Bench-Myopia~\cite{2023_benchmyopia_ebio} & \ding{55} & \ding{51} & \ding{55} & \ding{55} & \ding{55} & \ding{55} & \ding{55} & \ding{55} & \ding{55} \\
OphNet~\cite{2024_ophnet} & \ding{51} & \ding{51} & \ding{51} & \ding{55} & \ding{55} & \ding{55} & \ding{55} & \ding{55} & \ding{55} \\
\midrule 
\textbf{\dataset (ours)} & \ding{51} & \ding{51} & \ding{51} & \ding{51} & \ding{51} & \ding{51} & \ding{51} & \ding{51} & \ding{51} \\
\bottomrule
\end{tabular}
}
\label{tab:benchmark_comparison}
\end{table*}

This section examines recent developments in MLLMs and identifies critical gaps in comprehensive benchmarking resources for ophthalmic applications.

\subsection{Developments in Large Language Models (LLMs) and MLLMs}

\textbf{Evolution from BERT to Generative LLMs.} 
The past few years have seen a transformative shift in natural language processing with the advent of large-scale generative models. Early Transformer-based models like BERT~\cite{devlin2019bert} introduced bidirectional contextual understanding through masked language modeling, providing strong language representations that could be fine-tuned for diverse tasks~\cite{bosley2023we}. However, BERT-style models are not inherently generative and rely on task-specific fine-tuning, which limits their flexibility. In contrast, GPT-family models adopt an autoregressive learning objective – predicting the next token in a sequence – enabling open-ended text generation~\cite{bosley2023we}. This fundamental difference, combined with a dramatic increase in model scale (GPT-3~\cite{floridi2020gpt} contains 175 billion parameters versus BERT's 340 million~\cite{devlin2019bert}, endows LLMs with emergent capabilities for zero-shot and few-shot learning. For instance, GPT-3 demonstrated that even without explicit fine-tuning, a sufficiently large model can perform question-answering or summarization when prompted with a few examples. The release of ChatGPT in late 2022 further highlighted the potential of generative LLMs, as its instruction-tuned paradigm delivered human-like conversational abilities across a wide range of topics~\cite{tian2023opportunities}. Unlike earlier NLP systems, ChatGPT and its successors can engage in open-ended dialogue, generate detailed narratives, and adapt to user instructions in real-time, making them highly attractive for applications in education, healthcare, law, and beyond~\cite{bosley2023we}. This breakthrough has sparked extensive research into applying LLMs in specialized domains, such as answering biomedical questions, drafting clinical reports, or assisting with medical education. However, the deployment of LLMs in these high-stakes fields requires careful consideration of reliability and accuracy~\cite{tian2023opportunities}. Despite these remarkable text-based achievements, these models remained fundamentally limited to linguistic inputs, motivating researchers to explore multimodal extensions that could process visual information alongside natural language.

\textbf{Foundational MLLM Architectures and Early Explorations.}
Early pioneering systems explored different paradigms for integrating visual and textual modalities. OpenAI's CLIP~\cite{2021_clip_icml} aligned image representations with text embeddings through contrastive learning and showed that such alignment enables zero-shot image recognition via natural language prompts. DeepMind's Flamingo~\cite{alayrac2022flamingo} demonstrated an alternative architectural approach. It allows frozen pre-trained language models to effectively process visual inputs by integrating image features through gated cross-attention mechanisms (Perceiver Resampler). This design enables few-shot visual question answering without the need for vision-specific fine-tuning. This approach marked a significant breakthrough by showing that large language models could achieve visual reasoning capabilities through sophisticated architectural interfaces that dynamically fuse visual and textual information.

Since 2022, the field has witnessed explosive growth in MLLM development. Proprietary systems have led the charge in demonstrating advanced multimodal capabilities, though their architectural details remain largely undisclosed. OpenAI's GPT-4V~\cite{2023_openai_gpt4} demonstrates sophisticated multimodal reasoning, capable of interpreting complex images, analyzing diagrams, and even solving visual math problems without OCR, while Google's Gemini~\cite{team2023gemini} is reported to extend similar multimodal capabilities. These systems highlight the potential of large-scale multimodal training when supported by massive data and compute resources, and showcase emergent capabilities that were unattainable with earlier vision-language methods.

Building upon the foundational explorations and motivated by the success of these proprietary systems, the open-source community has developed a dominant architectural paradigm. Most modern MLLMs adopt a three-component architecture: (1) a visual encoder that extracts image representations, (2) a projection module that maps visual features into the LLM's input space, and (3) a language model that processes the combined multimodal inputs.
The visual encoder, commonly implemented as a convolutional neural network (CNN) or Vision Transformer (ViT)~\cite{dosovitskiy2020image}, generates a sequence of image feature embeddings from pre-trained visual representations.
A projection layer then transforms these visual embeddings to align with text by projecting them into the word embedding space.
Finally, the LLM, typically implemented as a decoder-only transformer, processes the projected visual features together with textual inputs to generate coherent multimodal outputs. This architectural design enables visual content to be effectively encoded as token-like representations that the language model can interpret and reason over.

While the three-component architecture (vision encoder–projector–LLM) provides a general framework, the specific choices in image preprocessing and cross-modal alignment mechanisms critically impact model capabilities. We detail these design decisions in representative MLLMs to illustrate current common practices.
LLaVA~\cite{2024_neurips_llava} adopts CLIP-style preprocessing, where images are resized and center-cropped to a fixed square resolution (e.g., 224--336\,px) and normalized using the CLIP statistics. A CLIP ViT encoder extracts patch embeddings, which are projected into the LLM’s word-embedding space through a lightweight two-layer MLP projector. This simple connector enables efficient end-to-end training of the vision--language interface.
InternVL~\cite{2024_internvl_cvpr} introduces dynamic high-resolution processing through adaptive tiling. Images are divided into a variable number of $448\times 448$ tiles (from a single tile up to a few dozen tiles depending on aspect ratio and resolution, supporting inputs up to 4K), and each tile is encoded by InternViT-6B. The resulting vision features are fed into a lightweight PixelShuffle-based downsampling layer that reduces spatial dimensions, before being concatenated with text tokens for joint processing by the LLM. This dynamic tiling strategy enables fine-grained recognition of text and detailed structures in high-resolution document or technical images.
Qwen-VL~\cite{bai2023qwen} uses a CLIP-style ViT encoder (e.g., OpenCLIP ViT-bigG) with images resized to a fixed square resolution (up to $448\times 448$) and split into patch tokens. To alleviate the long-sequence burden of feeding all visual patches into the LLM, Qwen-VL introduces a randomly initialized cross-attention--based vision--language adapter that compresses the dense visual tokens into a fixed-length sequence. The adapter employs fine-grained 2D absolute positional encodings to preserve spatial layout during this compression. The resulting compact visual tokens are then injected into the LLM’s token stream through designed image delimiters and prompts, enabling the model to perform grounding and multilingual text reading in images.
For our experimental evaluations, we directly utilize each model's native preprocessing pipeline without additional modifications. Images are processed and loaded according to the default configurations of each VLM.

The open-source community has rapidly advanced multimodal capabilities. LLaVA pioneered the influential encoder–projector–LLM framework that combines a CLIP vision encoder with LLaMA~\cite{touvron2023llama} via a linear projector for interactive image understanding. This foundational approach inspired numerous enhancements: BLIP-2~\cite{li2023blip} introduces a Q-Former, a lightweight Transformer adaptor with learnable queries that distills image features into compact tokens for frozen LLMs, with InstructBLIP~\cite{dai2024instructblip} further improving instruction following; MiniGPT-4~\cite{zhu2023minigpt} builds on BLIP-2 by training only a linear projector with minimal overhead; and VILA~\cite{2023_vila} leverages interleaved image–text pre-training to unlock in-context and multi-image reasoning with compact models.
However, these models primarily focused on low-to-moderate resolution image inputs (224×224 to 512×512 pixels), limiting fine-grained detail recognition and text readability in complex visual scenarios. Recent advances have addressed these constraints by supporting high-resolution images (up to 4K) and multi-modal inputs.
The InternVL~\cite{2024_internvl_cvpr} family exemplifies this evolution: InternVL 2.0 introduced dynamic tiling for high-resolution processing and extended to multi-image and video inputs, while InternVL 2.5 and the MPO variants further enhanced reasoning through improved training strategies and preference optimization. Similarly, DeepSeek-VL2~\cite{wu2024deepseekvl2mixtureofexpertsvisionlanguagemodels} achieves state-of-the-art performance through dynamic tiling and mixture-of-experts efficiency, while the Qwen-VL~\cite{bai2023qwen} family emphasizes multilingual understanding with multi-image interleaved inputs and region-level grounding capabilities. These high-resolution models enable precise OCR, detailed chart analysis, and complex multi-image reasoning that were previously unattainable in open-source systems.

\subsection{Gaps in Ophthalmology Datasets for the Development and Evaluation of LLMs and MLLMs}
Current datasets in ophthalmology are primarily designed for traditional supervised fine-tuning paradigms, typically constrained to single imaging modalities, specific tasks, or restricted output formats. In such settings, each model is fine-tuned for a predefined task with fixed input–output structures (e.g., disease severity levels)~\cite{khan2021global,2001_macularhole,ting2019artificial}. These datasets are well-suited for CNNs or vision transformers, which require task-specific fine-tuning, but are not feasible for LLMs and MLLMs. By contrast, LLMs and MLLMs possess zero-shot capabilities, allowing a single model to perform multiple tasks across diverse imaging modalities. Moreover, they are generative models that extend beyond fixed input–output mappings, enabling 
free-text generation which may provide thinking and 
reasoning steps~\cite{gilson2024language,zou2025benchmarking,yang2025towards}.

Pioneering efforts have introduced datasets for evaluating LLMs in ophthalmology~\cite{2024_benchnephrology_nejmai,2023_evalgptophth_ophsci,2023_benchmyopia_ebio,gilson2024language,srinivasan2025benchmarking}. However, most of these benchmarks remain language-only (e.g., ophthalmology knowledge testing), lacking ophthalmic imaging—arguably the most critical modality in clinical practice. Table~\ref{tab:benchmark_comparison} compares representative benchmarks for LLMs or MLLMs in both the general domain and ophthalmology across data modalities, imaging types, and applications. As shown, existing ophthalmology benchmarks primarily focus on single modalities or specific applications, making them insufficient for the development and evaluation of MLLMs.

More broadly, this reflects a significant gap in comprehensive benchmarks for AI development and evaluation in ophthalmology. A systematic review of 94 ophthalmology datasets~\cite{khan2021global} identified key limitations, including limited dataset scale, narrow task coverage, and the frequent absence of demographic information needed to assess potential performance biases. These limitations also raise concerns regarding the downstream accountability of AI in ophthalmology. Many studies report performance only on test sets that share distributions with their training data, often neglecting independent evaluations on external populations~\cite{liu2019comparison}.

\section{Method}

\begin{figure*}[tp]
    \centering
    \begin{subfigure}{0.95\textwidth}
        \centering
        \includegraphics[width=\linewidth]{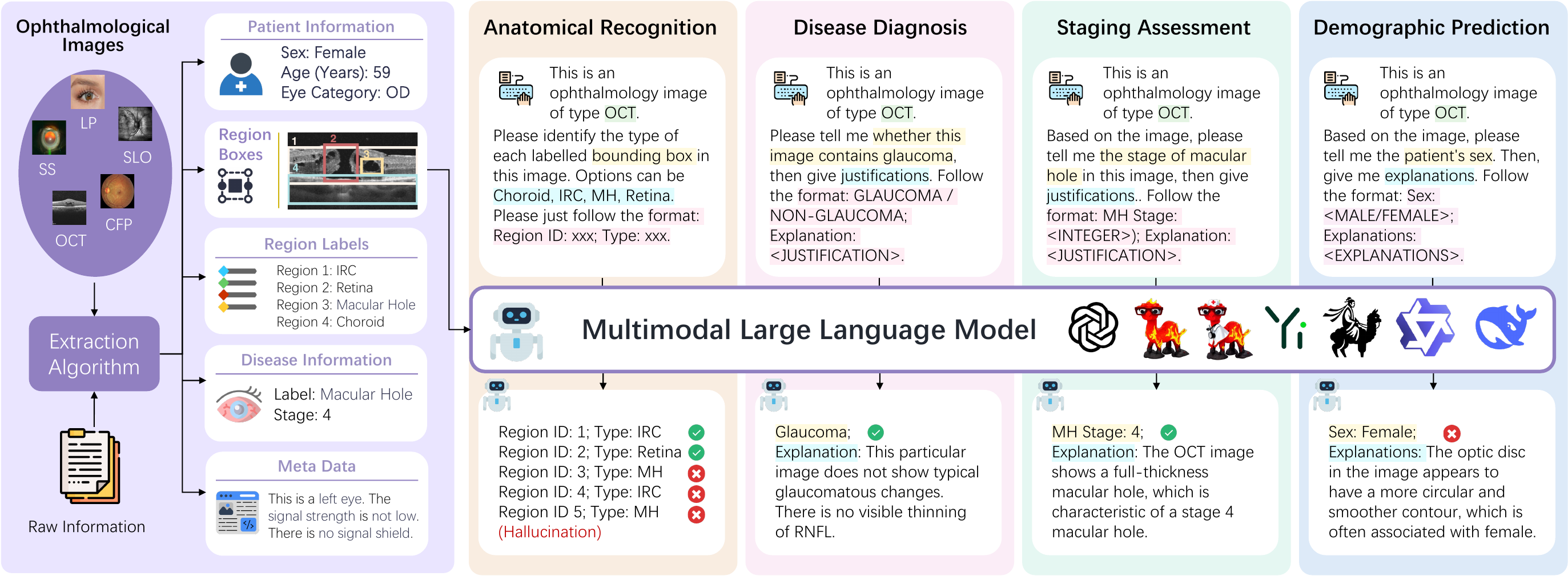}
        \caption{Data curation pipeline. We extract patient information, region bounding boxes and corresponding labels, disease information (including diagnosis and staging), and associated metadata. MLLMs are then employed to generate question–answer pairs that cover a wide range of ophthalmic tasks, including anatomical recognition, disease diagnosis, staging assessment, and demographic prediction.}
        \label{fig:data_pipeline}
    \end{subfigure}
    
    \vspace{1em}
    
    \begin{subfigure}{0.95\textwidth}
        \centering
        \includegraphics[width=\linewidth]{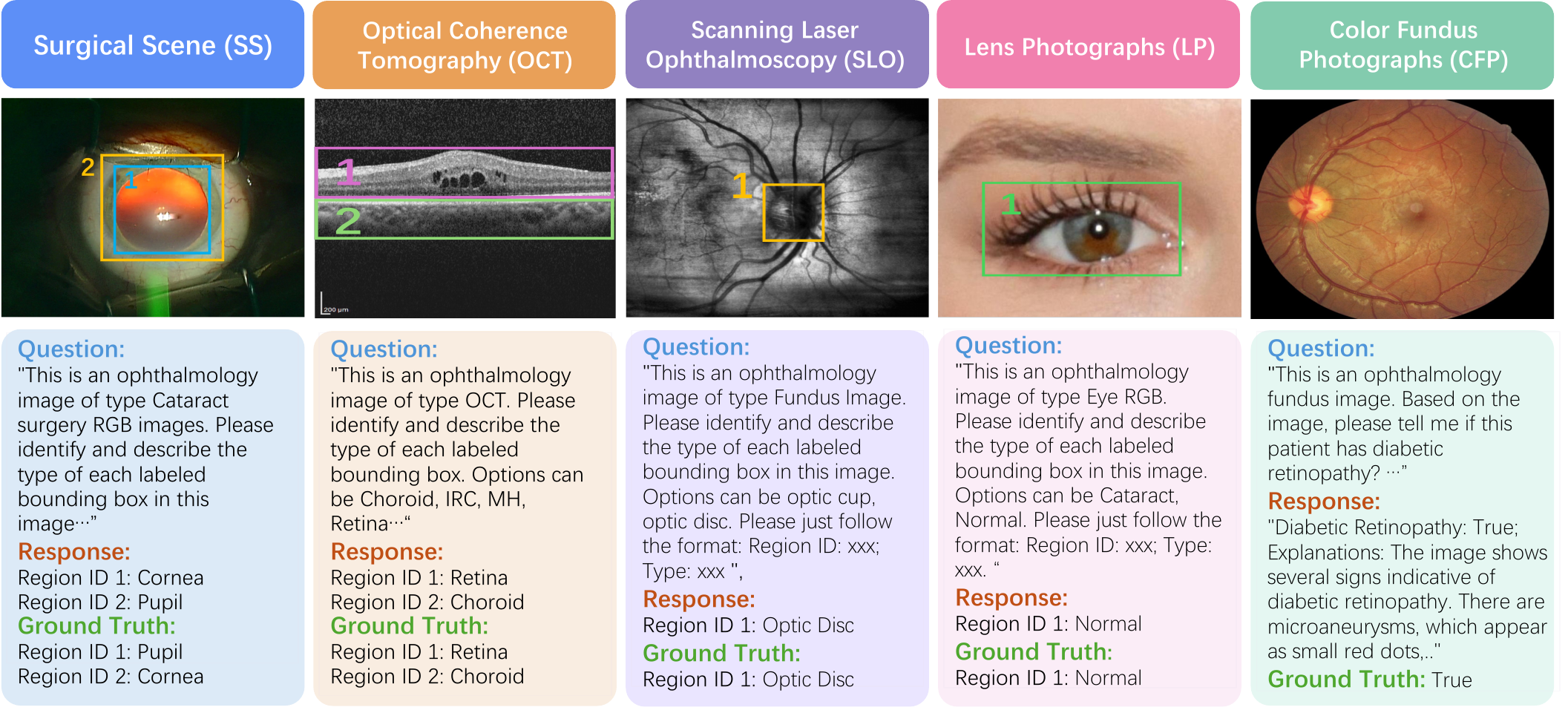}
        \caption{Detailed examples of the five ophthalmic imaging modalities included in our dataset. Surgical scene (SS), optical coherence tomography (OCT), scanning laser ophthalmoscopy (SLO), and lens photographs (LP) illustrate representative samples for anatomical recognition. Color fundus photography (CFP) demonstrates the binary eye condition diagnosis task.}
        \label{fig:data_examples}
    \end{subfigure}
    
    \caption{Overview of dataset construction and representative data samples: (a) data curation pipeline; (b) examples from multiple ophthalmic imaging modalities with corresponding task settings.}
    \label{fig:data_pipeline_examples}
    \Description{Two-part figure showing (a) a flowchart of the data curation pipeline that extracts patient information, bounding boxes, disease labels, and metadata from ophthalmic datasets, then uses MLLMs to generate question-answer pairs; and (b) representative image samples from five ophthalmic imaging modalities: surgical scenes, OCT, SLO, lens photographs, and color fundus photography, each with annotated bounding boxes and example prompts.}
\end{figure*}

As noted earlier, a primary challenge in advancing MLLMs for ophthalmology is the lack of unified, comprehensive benchmarks for development and evaluation. Most existing benchmarks were designed for earlier models such as CNNs, focusing on specific fine-tuning tasks and producing outputs as simple labels (e.g., presence or absence of AMD) rather than free-text responses. In this section, we present our data curation pipeline, which systematically repurposes existing benchmarks for MLLM development and evaluation, and describe subsequent systematic evaluations of 24 state-of-the-art MLLMs. The pipeline is publicly available and can be applied to emerging datasets and models.

\subsection*{Dataset Selection}

\textbf{Data Selection Criteria.} 
We selected representative ophthalmology datasets based on the following criteria: (1) they are publicly available with an open license or freely accessible for research use; (2) they are manually annotated by multiple domain experts; (3) they cover representative applications in ophthalmology, such as disease diagnosis and anatomical structure identification; and (4) they include demographic information, which is critical for generalization evaluation (e.g., across independent populations) and for assessing potential bias (e.g., prior studies have shown that AI models can predict gender from retinal photographs~\cite{korot2021predicting}).

\textbf{Data Sources and Composition.}
We scanned 20 publicly available datasets and selected 10 representative ones based on the criteria above. 
Collectively, these datasets span five distinct imaging modalities, as shown in Figure~\ref{fig:data_examples}, which represent key modalities in ophthalmology. This selection also facilitates a comprehensive evaluation of MLLM capabilities across the diverse imaging techniques used in contemporary ophthalmology. The datasets categorized by their imaging modalities
are detailed below.

\begin{itemize}
    \item \textit{Surgical Scene (SS)} imaging was represented by Cataract-1K~\cite{2024_cataract1k_scidata}, which contains 2,256 intraoperative images documenting cataract extraction procedures across multiple surgical phases.
    \item \textit{Optical Coherence Tomography (OCT)} included the OIMHS dataset~\cite{2023_oimhs_scidata}, comprising 3,859 macular OCT scans with expert-validated macular hole staging annotations. This modality provides high-resolution cross-sectional retinal imaging essential for detailed pathology assessment.
    \item \textit{Scanning Laser Ophthalmoscopy (SLO)} was represented by Harvard FairSeg~\cite{2024_harvardfairseg_tmi}, featuring 10,000 SLO fundus images with standardized optic disc and cup segmentations.
    \item \textit{Lens Photography (LP)} encompassed two complementary datasets. CAU001~\cite{2023_cau001} provided 1,417 anterior segment photographs of normal eyes, each annotated with the boundaries of the pupil, iris, and sclera. Cataract Detection 2~\cite{2023_cataract_det_2} contributed 1,015 lens photographs specifically designed for cataract detection.
    \item \textit{Color Fundus Photography (CFP)} constituted the largest component, incorporating multiple established datasets spanning major retinal pathologies. REFUGE~\cite{2020_refuge_miccaiw} contributed 1,200 images with validated glaucoma classifications and optic disc segmentations. IDRiD~\cite{2018_idrid_dataport} provided 516 images with pixel-level diabetic retinopathy lesion annotations for detailed pathology localization. ORIGA~\cite{2010_origa} contributed 650 images with comprehensive glaucoma measurements, while G1020~\cite{2020_g1020_ijcnn} added 1,020 high-resolution photographs with detailed glaucoma assessments. 
    Additional pathology-specific datasets from the BRSET~\cite{nakayama2024brset} collection encompassed a comprehensive range of retinal conditions including diabetic retinopathy with International Clinical Diabetic Retinopathy (ICDR) severity scale and Scottish Diabetic Retinopathy Grading (SDRG) scheme annotations, age-related macular degeneration, drusen, increased cup-to-disc ratio, vascular occlusions, myopic changes, hypertensive retinopathy, retinal hemorrhages, scarring, macular pathology, retinal nevi, and vascular structure annotations.
\end{itemize}

\begin{algorithm}[!t]
\captionsetup{font=small}
\caption{\scriptsize Anatomical Recognition Pipeline}
\label{alg:data_curation}
\begin{scriptsize}
\begin{algorithmic}[1]
\STATE \textbf{Input:} Original dataset \\ $D = \{(I_1, R_1), (I_2, R_2), \ldots, (I_n, R_n)\}$, where $I_i$ is an image and $R_i$ is the corresponding raw data
\STATE \textbf{Input:} Minimum bounding box area threshold $\tau \in \mathbb{R}^+$
\STATE \textbf{Output:} Curated dataset \\
$D' = \{(I_1, B'_1, P_1), (I_2, B'_2, P_2), \ldots, (I_n, B'_n, P_n)\}$, where $B'_i$ is the set of curated bounding boxes and $P_i$ is the set of corresponding prompts for image $I_i$
\FOR{each image-raw data pair $(I_i, R_i) \in D$}
    \STATE $B_i \leftarrow \text{ExtractBoundingBoxes}(R_i)$, \\
    where $B_i = \{b_{i,1}, b_{i,2}, \ldots, b_{i,|B_i|}\}$ and $b_{i,j}$ is the $j$-th bounding box of image $I_i$
\ENDFOR
\STATE $B \leftarrow \bigcup_{i=1}^n B_i$
\STATE $B' \leftarrow \{b \in B \mid \text{area}(b) \geq \tau\}$
\FOR{each image-raw data pair $(I_i, R_i) \in D$}
    \STATE $B'_i \leftarrow \{b \in B' \mid b \text{ belongs to image } I_i\}$
    \STATE $P_i \leftarrow \emptyset$
    \FOR{each bounding box $b_{i,j} \in B'_i$}
        \STATE $\textit{id}_{i,j} \leftarrow \text{GenerateUniqueID}()$
        \STATE $\textit{color}_{i,j} \leftarrow \text{AssignDistinctColor}()$
        \STATE $\textit{prompt}_{i,j} \leftarrow \text{GeneratePrompt}(b_{i,j})$
        \STATE $P_i \leftarrow P_i \cup \{(\textit{id}_{i,j}, \textit{color}_{i,j}, \textit{prompt}_{i,j})\}$
    \ENDFOR
\ENDFOR
\STATE \textbf{return} $D'$
\end{algorithmic}
\end{scriptsize}
\end{algorithm}

\subsection*{Evaluation Task Definition}
We further repurposed these datasets with a unified framework across primary ophthalmology applications~\cite{lu2018applications, ting2019artificial,wu2020application}, including: (1) anatomical structure recognition (identifying key anatomical components from images), (2) disease diagnosis (detecting the presence or absence of a single ophthalmic disease, or identifying which one among multiple ophthalmic diseases is present for screening), and (3) disease staging assessment (classifying disease severity). In addition, we further added an evaluation task to quantify whether MLLMs can predict demographic information from images for the assessment of potential bias. The tasks are described in detail below. 

\textit{Task 1: Anatomical Structure Recognition.}
Accurate recognition of localized ocular structures is essential for clinical imaging description and documentation (where clinicians manually summarize imaging findings), and it plays a critical role in supporting ophthalmic disease diagnosis~\cite{tong2020application,wu2020application}.
This task assesses the effectiveness of MLLMs to identify critical anatomical components
from the key ophthalmic imaging modalities described above.
Examples for this task can be found in SS, OCT, SLO and LP cases in Figure~\ref{fig:data_examples}.

\textit{Task 2: Ophthalmic Disease Diagnosis.}
This task evaluates MLLM diagnostic capabilities through two approaches: binary disease identification, where models determine the presence or absence of specific conditions, and multi-class disease diagnosis, where models determine which specific condition is present among multiple possible diseases, as illustrated in the CFP case in Figure~\ref{fig:data_examples}. These tasks assess fundamental diagnostic capabilities for common ophthalmic conditions and can be directly applied to screening and initial diagnostic workflows~\cite{ting2019artificial,mukherjee2025artificial}.

\textit{Task 3: Ophthalmic Disease Staging Assessment.}
In addition to the binary classification of eye disease conditions (present or absent), 
disease staging assessment further categorizes the severity levels of a condition. 
For instance, the International Clinical Diabetic Retinopathy (ICDR) scale 
(grades 0–4, ranging from no retinopathy to proliferative disease) is commonly used 
for staging diabetic retinopathy in clinical practice\footnote{\url{https://www.aao.org/education/clinical-statement/international-clinical-classification-system-diabe}}.
Disease staging is critical for monitoring progression and enabling early intervention~\cite{lu2018applications, chen2025ai}.

\textit{Task 4: Demographics Prediction.}
In addition to the key ophthalmology applications described above, this task evaluates the ability of MLLMs to infer patient demographic attributes, such as age and sex, directly from ocular imaging data. Prior studies have shown that AI models can predict gender from retinal photographs, raising concerns about potential bias if models rely primarily on demographic variables for inference~\cite{korot2021predicting, betzler2021gender}. We therefore included this task as an additional evaluation to assess potential bias.

We considered the data imbalance problem during evaluation and applied a class-balanced protocol for the three primary tasks: disease diagnosis, disease staging, and demographic prediction. For each task, we built a class-balanced evaluation subset via stratified subsampling and reported their accuracy as the primary metric. Accordingly, all results for these tasks were computed on the class-balanced evaluation sets. Detailed counts were provided in Table~\ref{tab:subsample_balanced_sizes}.

\subsection*{Data Curation Pipeline}
We developed a unified annotation pipeline to transform heterogeneous dataset formats into MLLM-compatible evaluation frameworks across the tasks.

\begin{table}[H]
\centering
\caption{Overview of anatomical structure recognition subset, including the number of images (Num Images) and average number of bounding boxes per image (Num Avg Boxes). }
\resizebox{0.5\textwidth}{!}{
\begin{tabular}{l r r}
\toprule
\textbf{Data Types} & \textbf{Num Images} & \textbf{Num Avg Boxes} \\
\midrule
Surgical Scenes (SS) & 2,256 & 3.3 \\
Optical Coherence Tomography (OCT) & 3,859 & 2.4 \\
Scanning Laser Ophthalmoscopy (SLO) & 10,000 & 1.0 \\
Lens Photography (LP) & 2,432 & 1.9 \\
Color Fundus Photography (CFP)  & 3,386 & 1.6 \\
\bottomrule
\end{tabular}
}
\label{tab:dataset_overview}
\end{table}
\textbf{Anatomical Structure Recognition. }
For datasets containing anatomical structure annotations such as segmentation masks or bounding box coordinates, we implemented a unified extraction and standardization process, as shown in Algorithm~\ref{alg:data_curation}. Raw annotations were converted into standardized bounding box coordinates and filtered using area-based thresholds to remove anatomically insignificant regions. Each structure received unique identifiers and distinct color codes for visual differentiation.
Automated prompt generation established correspondence between spatial annotations and natural language queries, enabling MLLM evaluation of anatomical recognition capabilities across imaging modalities. Representative prompts included: "This is an ophthalmology image of type Cataract surgery RGB images. Please identify and describe the type of each labeled bounding box in this image. Options can be Capsulorhexis Cystotome, Capsulorhexis Forceps, Cornea, Gauge, Incision Knife, Irrigation-Aspiration, Katena Forceps, Lens, Lens Injector, Phacoemulsification Tip, Pupil, Slit Knife, Spatula, cornea1. Please just follow the format: Region ID: xxx; Type: xxx."
The statistics of this subset can be found in Table~\ref{tab:dataset_overview}.

\textbf{Ophthalmic Disease Diagnosis. }
Image-level disease labels were transformed into structured prompt-response pairs suitable for MLLM evaluation. For binary condition identification, we employed prompts requiring definitive diagnostic decisions with explanatory rationale: ``This is an ophthalmology fundus image. Based on the image, please tell me if this patient has Age-Related Macular Degeneration (AMD)? Then, give me explanations. Follow the format: AMD \textless TRUE/FALSE\textgreater; Explanations: \textless EXPLANATIONS\textgreater.'' Multi-class diagnostic scenarios utilized comparative prompts: ``This is a colorful fundus image. Based on the image, please tell me the disease among cataract, diabetic retinopathy, glaucoma, normal. Then, give me explanations. Follow the format: DISEASE: \textless disease\_name\textgreater; Explanations: \textless EXPLANATIONS\textgreater.'' This approach mirrors clinical decision-making processes where physicians must justify diagnostic conclusions with supporting evidence.

\textbf{Ophthalmic Disease Staging Assessment. }
Ordinal staging labels were converted to prompt-based severity assessments reflecting clinical staging protocols. Representative prompts included: "This is an ophthalmology OCT image. Based on the image, please tell me the stage of <DISEASE> decision. Follow the format: Stage: <AN INTEGER>." This template evaluates MLLM capacity for fine-grained disease progression assessment, a critical capability distinguishing experienced clinicians who can discern subtle morphological changes indicative of disease advancement.

\textbf{Patient Demographic Attribute Prediction. }
Patient-level demographic labels (sex and age group) were reformulated into structured prompt–response templates for MLLM evaluation. For binary sex identification, we employed prompts requiring categorical decisions with explanatory justification: ``This is an ophthalmology fundus image. Based on the image, please tell me the patient's sex. Then, give me explanations. Follow the format: Sex: \textless MALE/FEMALE\textgreater; Explanations: \textless EXPLANATIONS\textgreater.'' For age group prediction, pre-defined categorical ranges were explicitly embedded within the prompt: ``This is an ophthalmology fundus image. Based on the image, please tell me the patient's age group. Then, give me explanations. The age groups are: Group 1: \textless18; Group 2: 18--29; Group 3: 30--39; Group 4: 40--49; Group 5: 50--59; Group 6: 60--69; Group 7: 70--79; Group 8: 80+. Follow the format: Age Group: \textless GROUP\_LABEL\textgreater; Explanations: \textless EXPLANATIONS\textgreater.'' While the model was prompted with these fine-grained categories, for evaluation we further consolidated the predictions into 4 broader groups — 18–40, 40–60, 60+, and Invalid — following established medical and public health standards from the National Center for Health Statistics \cite{ostchega2020}. This design enables systematic evaluation of MLLM capacity to infer demographic characteristics from ocular imaging
data, thereby assessing MLLM potential bias in demographic inference.

To facilitate reproducible and quick benchmarking, we released a carefully curated balanced subset used only at evaluation time.
It comprised four-class disease classification with 100 samples per class, four-stage disease staging with 19 samples per stage, and anatomy drawn from three sources: CAU001, Cataract\textendash1K, and ORIGA, with 200 samples from each source.
Full counts are provided in Table~\ref{tab:balanced_subset_small}, and results based on this subset are reported on the project webpage.

\subsection*{Systematic Evaluation}
We systematically evaluated the effectiveness of 24 representative MLLMs on the benchmark. For each task, we employed commonly used metrics and incorporated additional measures tailored to generative models, such as hallucination related measures. The evaluation metrics are detailed below.

\textbf{Evaluation metrics.}
For anatomical structure recognition, we employed a comprehensive set of metrics to evaluate model performance in identifying and localizing ophthalmic anatomical features.

\textbf{Precision}: Measures the proportion of correctly predicted region types among all predicted regions. A high precision indicates that the model is more likely to be correct when predicting region types. 

\textbf{Recall}: Quantifies the proportion of correctly predicted region types among all ground truth regions. A high recall indicates that the model is able to identify a larger fraction of the relevant regions.

\textbf{F1 Score}: The harmonic mean of precision and recall, providing a balanced measure.

\textbf{Hallucination Resistance (HR):} The Hallucination Resistance (HR) metric is proposed by us to quantify a model's ability to avoid hallucinations: 

\begin{equation*} 
\label{eq:hr_metric_cardinality}
\text{HR} = 1 - \frac{|\{r \in \mathcal{P}_i \mid r \notin \mathcal{T}_i\}|}{|\{r \in \mathcal{P}_i \}|},
\end{equation*}

where $\mathcal{P}_i$ represents set of all predicted region IDs for image $i$, and
$\mathcal{T}_i$ indicates set of all ground truth region IDs. Higher HR values indicate fewer hallucinations.

In addition, we used accuracy as the primary evaluation metric for the other tasks, as we ensured the datasets are balanced at the evaluation stage. For ophthalmic disease diagnosis, we reported both binary accuracy (for the classification of a single eye condition) and multi-class accuracy (for detecting disease among multiple diseases). For disease staging, we reported overall accuracy.
For demographic prediction, since age is a continuous variable, we grouped ages into categories and used accuracy as the evaluation measure.

\textbf{Model representatives.}
We evaluated 24 state-of-the-art MLLMs selected from different perspectives for comprehensive coverage. 
General-purpose models included the closed-source GPT-4o~\cite{2023_openai_gpt4} and representative open-weight models such as Yi-VL-6B~\cite{2024_yi}, the LLaVA series~\cite{2024_neurips_llava}, Qwen series~\cite{bai2023qwen}, InternVL series~\cite{2024_internvl_cvpr} with mixed preference optimization (MPO) variants~\cite{wang2025enhancingreasoningabilitymultimodal}, and the DeepSeek-VL series~\cite{wu2024deepseekvl2mixtureofexpertsvisionlanguagemodels}. 
These models have been widely adopted in the general domain and consistently report state-of-the-art performance across diverse multimodal tasks~\cite{liang2024survey,li2025survey}. In addition, we also included medical-specific models such as LLaVA-Med~\cite{2024_neurips_llavamed} and Med-Flamingo~\cite{2022_flamingo_neurips}, which represent pioneering efforts to adapt MLLMs for medical applications.

\section{Results}
\begin{table*}[t]
\centering
\caption{Overall performance of 24 MLLMs on the LMOD benchmark, reported as weighted averages across four tasks. The ``Random'' baseline samples answers uniformly at random. Demographics prediction is included to assess potential bias, with age grouped into four categories: 18--40, 40--60, 60+, and ``Invalid'' (missing or inconsistent data).}

\label{tab:main_table_results}
\normalsize
\setlength{\tabcolsep}{3pt}
\renewcommand{\arraystretch}{1.2}
\resizebox{\textwidth}{!}{
\begin{tabular}{lccccccc|ccc}
\toprule
\multirow{2}{*}{\textbf{Models}} 
& \multicolumn{4}{c}{\textbf{Anatomical Recognition}} 
& \multicolumn{2}{c}{\textbf{Diagnosis Analysis}} 
& \multicolumn{1}{c}{\textbf{Staging Assessment}} 
& \multicolumn{2}{|c}{\textbf{Demographics Prediction}} \\
\cmidrule(lr){2-5} \cmidrule(lr){6-7} \cmidrule(lr){8-8} \cmidrule(lr){9-10}
& Prec. & Rec. & F1 & HR 
& Binary Acc & Multi-class Acc
& Acc
& Sex Acc & Age Acc \\ 
\midrule
Random & - & - & - & - & 0.5000 & 0.2500 & 0.2393 & 0.5000 & 0.2500 \\
\midrule
GPT-4o & 0.5807 & \textbf{0.5766} & \textbf{0.5761} & 0.9439 & - & - & 0.1971 & - & - \\
LLaVA-Med-v1.5-mistral-7B & 0.0789 & 0.1163 & 0.0789 & 0.7434 & 0.3882 & \textbf{0.3626} & 0.2453 & 0.5000 & 0.2500 \\
Yi-VL-6B & 0.1948 & 0.1495 & 0.1615 & 0.8480 & 0.4968 & 0.2763 & 0.2486 & 0.5000 & 0.2538 \\
Med-Flamingo & - & - & - & - & - & - & - & - & - \\
\midrule
\noalign{\vskip -3pt}
\multicolumn{10}{c}{\textbf{InternVL Series}} \\
\noalign{\vskip -3pt}
\midrule
InternVL-1.5-2B & \underline{0.6026} & 0.3999 & 0.4630 & 0.9807 & 0.4993 & 0.2500 & \underline{0.2587} & 0.5000 & 0.2555 \\
InternVL-1.5-4B & \textbf{0.7249} & \underline{0.4996} & \underline{0.5716} & 0.9624 & 0.5267 & 0.2575 & 0.2556 & 0.5000 & 0.2500 \\
InternVL-2.0-2B & 0.0954 & 0.1100 & 0.0836 & 0.7948 & 0.4803 & 0.2498 & 0.2407 & 0.5000 & 0.2511 \\
InternVL-2.0-4B & 0.3609 & 0.2456 & 0.2353 & 0.8387 & 0.5251 & 0.2204 & 0.2082 & 0.5059 & 0.2500 \\
InternVL-2.0-8B & 0.4214 & 0.3232 & 0.3168 & 0.9406 & 0.5570 & \underline{0.3617} & 0.2466 & 0.5004 & 0.2509 \\
InternVL-2.5-2B & 0.1116 & 0.1144 & 0.0994 & 0.8793 & 0.5569 & 0.2952 & 0.0464 & 0.5000 & \underline{0.3076} \\
InternVL-2.5-4B & 0.2614 & 0.1662 & 0.1715 & 0.9828 & 0.5339 & 0.3309 & 0.2427 & 0.0000 & 0.2561 \\
InternVL-2.5-8B & 0.4672 & 0.4061 & 0.4031 & 0.9789 & \underline{0.5783} & 0.3595 & \textbf{0.2667} & \underline{0.5069} & 0.2367 \\
InternVL-2.5-2B-MPO & 0.0525 & 0.0620 & 0.0497 & 0.8794 & 0.5131 & 0.2530 & 0.2442 & 0.5000 & 0.2661 \\
InternVL-2.5-4B-MPO & 0.2890 & 0.1649 & 0.1764 & \textbf{0.9943} & 0.5713 & 0.3473 & 0.2433 & 0.0000 & 0.2519 \\
InternVL-2.5-8B-MPO & 0.4411 & 0.3494 & 0.3545 & 0.9819 & 0.5612 & 0.3538 & 0.2084 & 0.0000 & 0.2965 \\
\midrule
\noalign{\vskip -3pt}
\multicolumn{10}{c}{\textbf{LLaVA Series}} \\
\noalign{\vskip -3pt}
\midrule
LLaVA-1.5-7B &  0.0567 & 0.0410 & 0.0456 & 0.2675 & 0.5056 & 0.2461 & 0.2391 & \textbf{0.5105} & 0.2463 \\
LLaVA-Mistral-7B & 0.1274 & 0.1503 & 0.1285 & 0.5676 & 0.5033 & 0.2547 & 0.2353 & 0.5000 & 0.2778 \\
LLaVA-Vicuna-7B & 0.3086 & 0.2534 & 0.2668 & 0.7105 & 0.4857 & 0.2807 & 0.1868 & 0.5000 & 0.2350 \\
LLaVA-Vicuna-13B & 0.0544 & 0.0730 & 0.0591 & 0.3731 & 0.5028 & 0.2224 & 0.2148 & 0.4993 & 0.1883 \\
\midrule
\noalign{\vskip -3pt}
\multicolumn{10}{c}{\textbf{Qwen Series}} \\
\noalign{\vskip -3pt}
\midrule
Qwen-VL-Chat & 0.0270 & 0.0365 & 0.0274 & 0.8398 & 0.4966 & 0.2561 & 0.2360 & 0.5000 & \textbf{0.3457} \\
Qwen-3B & 0.3576 & 0.2038 & 0.2238 & 0.7241 & 0.5229 & 0.2599 & 0.2527 & 0.5014 & 0.2500 \\
Qwen-7B & 0.2614 & 0.1704 & 0.1814 & 0.7079 & \textbf{0.5826} & 0.2459 & 0.2409 & 0.4999 & 0.2517 \\
\midrule
\noalign{\vskip -3pt}
\multicolumn{10}{c}{\textbf{DeepSeek Series}} \\
\noalign{\vskip -3pt}
\midrule
DeepSeek-VL2-Tiny & 0.2110 & 0.1738 & 0.1796 & \underline{0.9891} & 0.5030 & 0.2604 & 0.0842 & 0.4975 & 0.2500 \\
DeepSeek-VL2-Small & 0.0211 & 0.0035 & 0.0055 & 0.4433 & - & 0.2665 & 0.0425 & - & 0.2050 \\
\midrule
Average & 0.2656 & 0.2082 & 0.2113 & 0.7988 & 0.5186 & 0.2823 & 0.2124 & 0.4296 & 0.2446 \\
\bottomrule
\addlinespace[2pt]
\multicolumn{10}{l}{\normalsize \textit{Note:} \textbf{Bold} indicates the best performance in each column; \underline{underline} indicates the second best; ``--'' denotes inapplicable results} \\
\end{tabular}
}
\end{table*}

Table~\ref{tab:main_table_results} presents an overview of the performance of all 24 models across tasks. Detailed results for each individual task are summarized below.

\subsection*{Anatomical Structure Recognition}

\textbf{Overall Performance.} As shown in Table~\ref{tab:main_table_results}, GPT-4o consistently achieved superior performance across all evaluation metrics and obtained the highest F1 score (57.61\%). Notably, the open-weight InternVL-1.5-4B demonstrated highly competitive results (F1 = 57.16\%) relative to GPT-4o, while also exhibiting stronger resistance to hallucinations. In contrast, several medical MLLMs, such as Med-Flamingo and LLaVA-Med, performed substantially worse (e.g., F1 $\sim$7\%), suggesting that in this context, medical MLLMs did not necessarily demonstrate improved performance on medical specialties. Moreover, considerable performance variance was observed among the 24 MLLMs, even within the same family and parameter size, indicating instability in model generalization. Overall, despite recent progress, none of those models achieved satisfactory performance in anatomical recognition in ophthalmology, suggesting a
pressing need to develop domain-specific models.

\begin{figure*}[h!]
    \centering
    \includegraphics[width=\textwidth]{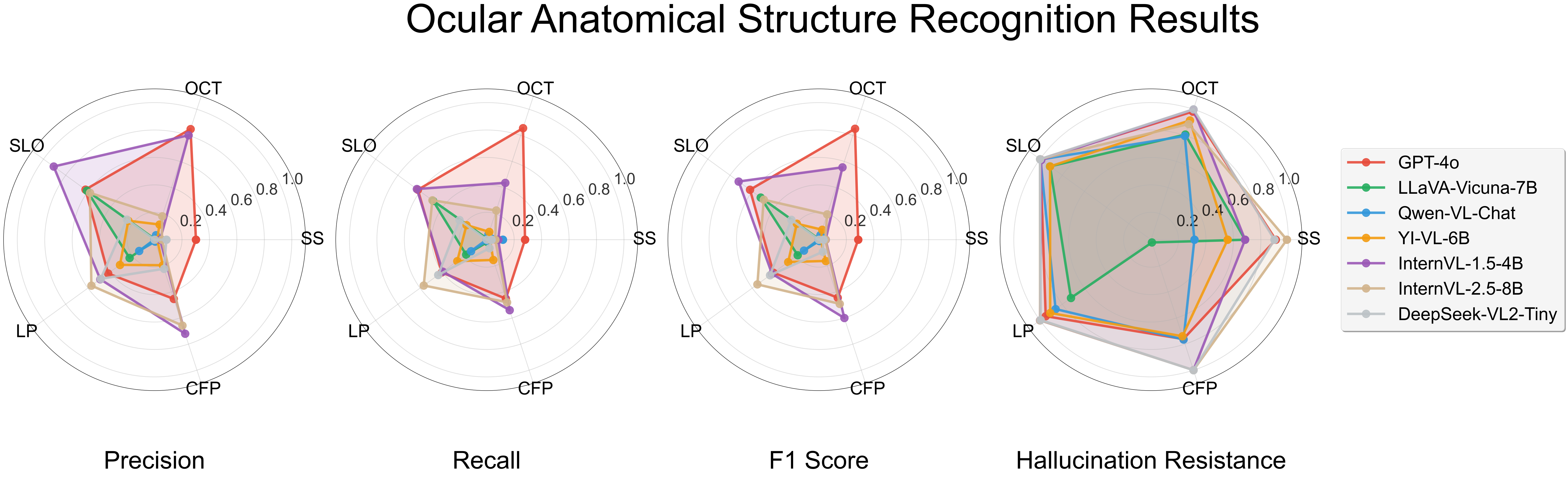}
    \caption{Performance comparison of top-performing MLLMs across different ophthalmic imaging modalities. The radar charts display the performance of the top-F1-performing models, for each evaluation metric (Precision, Recall, F1, and HR) across five different imaging modalities: surgical scenes (SS), optical coherence tomography (OCT), color fundus photographs (CFP), scanning laser ophthalmoscopy (SLO), and lens photographs (LP). }
    \label{fig:radar_chart}
    \Description{Radar charts comparing top-performing MLLMs across five ophthalmic imaging modalities (SS, OCT, CFP, SLO, LP) on four metrics (Precision, Recall, F1, HR), showing that model performance varies substantially across modalities.}
\end{figure*}

\begin{table*}[t]
\centering
\caption{Anatomical structure recognition results of 24 MLLMs on five ophthalmic imaging modalities, split into two side-by-side subtables (the right subtable continues methods from the left). GPT-4o consistently achieved superior performance across all evaluation metrics, while the open-weight InternVL-1.5-4B demonstrated highly competitive results relative to GPT-4o.}
\label{tab:anatomical_recognition_processed_split}
\setlength{\tabcolsep}{4pt}
\renewcommand{\arraystretch}{1.1}
\newlength{\splitheight}
\setlength{\splitheight}{0.69\textheight}

\newcommand{\hdr}[1]{\textbf{\Large #1}\rule{0pt}{3.2ex}\rule[-1.2ex]{0pt}{0pt}}

\begin{minipage}[t][\splitheight]{0.5\textwidth}
\centering
\resizebox{\linewidth}{!}{
\begin{tabular}{l|c|c|c|c|c|c|c}
\toprule
\hdr{Method} & \hdr{Metrics} & \hdr{Overall} & \hdr{SS} & \hdr{OCT} & \hdr{SLO} & \hdr{LP} & \hdr{CFP} \\
\midrule
\multirow{4}{*}{GPT-4o} & Precision & 0.5807 & \textbf{0.3017} & \underline{0.8484} & 0.6226 & 0.4186 & 0.4539 \\
 & Recall & \textbf{0.5766} & \textbf{0.2790} & \textbf{0.8555} & \underline{0.6199} & 0.4053 & 0.4520 \\
 & F1 & \textbf{0.5761} & \textbf{0.2864} & \textbf{0.8512} & \underline{0.6205} & 0.4093 & 0.4446 \\
 & HR & 0.9439 & 0.9085 & 0.9829 & 0.9974 & 0.9486 & 0.7616 \\
\midrule
\multirow{4}{*}{LLaVA-1.5-7B} & Precision & 0.0567 & 0.0145 & 0.0030 & 0.0700 & 0.0999 & 0.0759 \\
 & Recall & 0.0410 & 0.0078 & 0.0036 & 0.0504 & 0.0649 & 0.0611 \\
 & F1 & 0.0456 & 0.0076 & 0.0024 & 0.0570 & 0.0743 & 0.0661 \\
 & HR & 0.2675 & 0.4547 & 0.4196 & 0.1377 & 0.3870 & 0.2668 \\
\midrule
\multirow{4}{*}{LLaVA-Mistral-7B} & Precision & 0.1274 & 0.0144 & 0.1856 & 0.1011 & 0.2532 & 0.1236 \\
 & Recall & 0.1503 & 0.0710 & 0.1586 & 0.1455 & 0.2656 & 0.1250 \\
 & F1 & 0.1285 & 0.0231 & 0.1622 & 0.1122 & 0.2503 & 0.1210 \\
 & HR & 0.5676 & 0.2485 & 0.7415 & 0.4404 & 0.8635 & 0.7454 \\
\midrule
\multirow{4}{*}{LLaVA-Vicuna-7B} & Precision & 0.3086 & 0.0410 & 0.0047 & 0.6093 & 0.2253 & 0.0047 \\
 & Recall & 0.2534 & 0.0837 & 0.0085 & 0.4826 & 0.1851 & 0.0175 \\
 & F1 & 0.2668 & 0.0492 & 0.0042 & 0.5242 & 0.1910 & 0.0051 \\
 & HR & 0.7105 & 0.6851 & 0.8064 & 0.9101 & 0.7215 & 0.0208 \\
\midrule
\multirow{4}{*}{LLaVA-Vicuna-13B} & Precision & 0.0544 & 0.0085 & 0.0488 & 0.0704 & 0.0856 & 0.0220 \\
 & Recall & 0.0730 & 0.0217 & 0.0672 & 0.0896 & 0.1075 & 0.0398 \\
 & F1 & 0.0591 & 0.0099 & 0.0547 & 0.0751 & 0.0913 & 0.0268 \\
 & HR & 0.3731 & 0.1648 & 0.5538 & 0.3832 & 0.5003 & 0.1848 \\
\midrule
\multirow{4}{*}{LLaVA-Med-v1.5-mistral-7B} & Precision & 0.0789 & 0.0278 & 0.0326 & 0 & 0.1168 & 0.3715 \\
 & Recall & 0.1163 & \underline{0.1362} & 0.0397 & 0 & 0.3206 & 0.3872 \\
 & F1 & 0.0789 & 0.0462 & 0.0356 & 0 & 0.1332 & 0.3440 \\
 & HR & 0.7434 & 0.2220 & 0.6209 & 0.9997 & 0.3923 & 0.7257 \\
\midrule
\multirow{4}{*}{Qwen-VL-Chat} & Precision & 0.0270 & 0.0243 & 0.0345 & 0.0020 & 0.1383 & 0.0139 \\
 & Recall & 0.0365 & 0.1174 & 0.0412 & 0.0011 & 0.1411 & 0.0070 \\
 & F1 & 0.0274 & 0.0399 & 0.0358 & 0.0014 & 0.1349 & 0.0093 \\
 & HR & 0.8398 & 0.3156 & 0.7954 & 0.9956 & 0.8600 & 0.7651 \\
\midrule
\multirow{4}{*}{Yi-VL-6B} & Precision & 0.1948 & 0.0198 & 0.1168 & 0.2358 & 0.3115 & 0.1955 \\
 & Recall & 0.1495 & 0.0246 & 0.0605 & 0.1819 & 0.2657 & 0.1549 \\
 & F1 & 0.1615 & 0.0146 & 0.0769 & 0.1989 & 0.2767 & 0.1626 \\
 & HR & 0.8480 & 0.5578 & 0.9151 & 0.9096 & 0.9085 & 0.7397 \\
\midrule
\multirow{4}{*}{Med-Flamingo} & Precision & -- & -- & -- & -- & -- & -- \\
 & Recall & -- & -- & -- & -- & -- & -- \\
 & F1 & -- & -- & -- & -- & -- & -- \\
 & HR & -- & -- & -- & -- & -- & -- \\
\midrule
\multirow{4}{*}{InternVL-1.5-2B} & Precision & \underline{0.6026} & 0.0367 & \textbf{0.8790} & 0.6428 & 0.3886 & \underline{0.6996} \\
 & Recall & 0.3999 & 0.0192 & 0.3999 & 0.4792 & 0.2523 & \underline{0.5254} \\
 & F1 & 0.4630 & 0.0208 & 0.5436 & 0.5337 & 0.2873 & \underline{0.5835} \\
 & HR & 0.9807 & 0.8163 & \underline{0.9992} & 0.9996 & \textbf{1.0000} & 0.9993 \\
\midrule
\multirow{4}{*}{InternVL-1.5-4B} & Precision & \textbf{0.7249} & 0.0409 & 0.8011 & \textbf{0.9085} & 0.4893 & \textbf{0.7206} \\
 & Recall & \underline{0.4996} & 0.0837 & \underline{0.4364} & \textbf{0.6295} & 0.3953 & \textbf{0.5397} \\
 & F1 & \underline{0.5716} & 0.0492 & \underline{0.5552} & \textbf{0.7225} & 0.4225 & \textbf{0.6000} \\
 & HR & 0.9624 & 0.6851 & 0.9989 & 0.9893 & \underline{0.9997} & 0.9994 \\
\midrule
\multirow{4}{*}{InternVL-2.0-2B} & Precision & 0.0954 & 0.0753 & 0.0407 & 0.0025 & 0.2679 & 0.3215 \\
 & Recall & 0.1100 & 0.0361 & 0.0798 & 0.0020 & 0.3773 & 0.3206 \\
 & F1 & 0.0836 & 0.0277 & 0.0535 & 0.0022 & 0.2984 & 0.2410 \\
 & HR & 0.7948 & 0.8551 & 0.6103 & 0.8539 & 0.7720 & 0.8070 \\
\bottomrule
\end{tabular}
}
\vfill
\end{minipage}\hfill
\begin{minipage}[t][\splitheight]{0.472\textwidth}
\centering
\resizebox{\linewidth}{!}{
\begin{tabular}{l|c|c|c|c|c|c|c}
\toprule
\hdr{Method} & \hdr{Metrics} & \hdr{Overall} & \hdr{SS} & \hdr{OCT} & \hdr{SLO} & \hdr{LP} & \hdr{CFP} \\
\midrule
\multirow{4}{*}{InternVL-2.0-4B} & Precision & 0.3609 & 0.0204 & 0.1580 & 0.5000 & 0.2585 & 0.4820 \\
 & Recall & 0.2456 & 0.0371 & 0.2203 & 0.2375 & 0.3912 & 0.3328 \\
 & F1 & 0.2353 & 0.0175 & 0.1766 & 0.2596 & 0.2627 & 0.3558 \\
 & HR & 0.8387 & 0.7217 & 0.5949 & 0.9394 & 0.8823 & 0.8659 \\
\midrule
\multirow{4}{*}{InternVL-2.0-8B} & Precision & 0.4214 & 0.0465 & 0.1272 & 0.5271 & \underline{0.5653} & 0.5909 \\
 & Recall & 0.3232 & 0.0582 & 0.1711 & 0.3321 & \underline{0.5303} & 0.4983 \\
 & F1 & 0.3168 & 0.0477 & 0.1373 & 0.3339 & \underline{0.5180} & 0.5055 \\
 & HR & 0.9406 & 0.9805 & 0.6790 & 0.9999 & 0.9958 & 0.9976 \\
\midrule
\multirow{4}{*}{InternVL-2.5-2B} & Precision & 0.1116 & 0.1135 & 0.0182 & 0.0274 & 0.3763 & 0.2756 \\
 & Recall & 0.1144 & 0.0540 & 0.0209 & 0.0290 & 0.3557 & 0.3402 \\
 & F1 & 0.0994 & 0.0479 & 0.0173 & 0.0269 & 0.3269 & 0.2783 \\
 & HR & 0.8793 & 0.9378 & 0.9303 & 0.8739 & 0.9257 & 0.7650 \\
\midrule
\multirow{4}{*}{InternVL-2.5-4B} & Precision & 0.2614 & 0.0466 & 0.0848 & 0.2629 & 0.5527 & 0.3923 \\
 & Recall & 0.1662 & \textbf{0.9986} & 0.0862 & 0.1009 & 0.5205 & 0.2556 \\
 & F1 & 0.1715 & 0.0469 & 0.0789 & 0.1179 & 0.4800 & 0.2966 \\
 & HR & 0.9827 & \textbf{0.9986} & 0.9086 & 0.9997 & 0.9988 & 0.9952 \\
\midrule
\multirow{4}{*}{InternVL-2.5-8B} & Precision & 0.4671 & 0.0539 & 0.1804 & 0.5819 & \textbf{0.5690} & 0.6575 \\
 & Recall & 0.4061 & 0.0623 & 0.2243 & 0.4898 & \textbf{0.5674} & 0.4794 \\
 & F1 & 0.4030 & 0.0469 & 0.1950 & 0.4971 & \textbf{0.5538} & 0.4916 \\
 & HR & 0.9788 & 0.9897 & 0.8868 & \textbf{1.0000} & 0.9991 & \textbf{0.9997} \\
\midrule
\multirow{4}{*}{InternVL-2.5-2B-MPO} & Precision & 0.0525 & 0.0245 & 0.0283 & 0.0051 & 0.0768 & 0.2216 \\
 & Recall & 0.0620 & 0.0374 & 0.0242 & 0.0055 & 0.0705 & 0.2820 \\
 & F1 & 0.0497 & 0.0272 & 0.0224 & 0.0051 & 0.0548 & 0.2237 \\
 & HR & 0.8794 & 0.9355 & 0.9070 & 0.8667 & 0.9705 & 0.7824 \\
\midrule
\multirow{4}{*}{InternVL-2.5-4B-MPO} & Precision & 0.2890 & 0.0774 & 0.1129 & 0.2898 & 0.5242 & 0.4594 \\
 & Recall & 0.1649 & 0.0888 & 0.0918 & 0.1044 & 0.4804 & 0.2507 \\
 & F1 & 0.1764 & \underline{0.0652} & 0.0876 & 0.1256 & 0.4542 & 0.3023 \\
 & HR & \textbf{0.9943} & \underline{0.9981} & 0.9720 & 0.9997 & 0.9988 & 0.9982 \\
\midrule
\multirow{4}{*}{InternVL-2.5-8B-MPO} & Precision & 0.4411 & 0.0681 & 0.0885 & \underline{0.6804} & 0.2559 & 0.5178 \\
 & Recall & 0.3494 & 0.0796 & 0.0846 & \underline{0.4942} & 0.2710 & 0.4598 \\
 & F1 & 0.3545 & 0.0647 & 0.0856 & \underline{0.5115} & 0.2407 & 0.4720 \\
 & HR & 0.9818 & 0.9946 & 0.9019 & \underline{1.0000} & 0.9974 & \underline{0.9997} \\
\midrule
\multirow{4}{*}{Qwen-3B} & Precision & 0.3576 & 0.0282 & 0.0599 & 0.4975 & 0.4916 & 0.4070 \\
 & Recall & 0.2038 & 0.0320 & 0.0956 & 0.2362 & 0.4204 & 0.1904 \\
 & F1 & 0.2238 & 0.0054 & 0.0693 & 0.3158 & 0.3445 & 0.1873 \\
 & HR & 0.7241 & 0.2361 & 0.5854 & 0.7883 & 0.9821 & 0.8323 \\
\midrule
\multirow{4}{*}{Qwen-7B} & Precision & 0.2614 & 0.0291 & 0.1283 & 0.3333 & 0.4191 & 0.2426 \\
 & Recall & 0.1704 & 0.0462 & 0.1591 & 0.1124 & 0.4581 & 0.2307 \\
 & F1 & 0.1814 & 0.0272 & 0.1300 & 0.1677 & 0.4051 & 0.2224 \\
 & HR & 0.7079 & 0.4141 & 0.4576 & 0.7784 & 0.9127 & 0.8337 \\
\midrule
\multirow{4}{*}{DeepSeek-VL2-Tiny} & Precision & 0.2110 & 0.0862 & 0 & 0.2500 & 0.4849 & 0.2227 \\
 & Recall & 0.1738 & 0.0346 & 0 & 0.2451 & 0.4370 & 0.0653 \\
 & F1 & 0.1796 & 0.0334 & 0 & 0.2479 & 0.4406 & 0.0923 \\
 & HR & \underline{0.9890} & 0.8992 & \textbf{1.0000} & 1.0000 & 0.9948 & 1.0000 \\
\midrule
\multirow{4}{*}{DeepSeek-VL2-Small} & Precision & 0.0211 & 0 & 0 & 0 & 0.1907 & 0.0000 \\
 & Recall & 0.0035 & 0.0000 & 0 & 0 & 0.0313 & 0.0000 \\
 & F1 & 0.0055 & 0 & 0 & 0 & 0.0498 & 0.0000 \\
 & HR & 0.4433 & 0 & 0.7273 & 0.3333 & 0.9947 & 0.3441 \\
\bottomrule
\end{tabular}
}
\vfill
\end{minipage}

\parbox{\textwidth}{\footnotesize \textit{Note:} \textbf{Bold} indicates the best performance among all the methods; \underline{underline} indicates the second best; ``--'' denotes inapplicable results.}
\end{table*}
\textbf{Model Family Analysis.} 
Figure~\ref{fig:radar_chart} shows anatomical structure recognition results across imaging modalities, with detailed results provided in Table~\ref{tab:anatomical_recognition_processed_split}. We compared performance by model family. GPT-4o achieved the most consistent results across all five ophthalmic imaging modalities, with particularly strong performance in OCT interpretation (F1 = 0.8512). The InternVL family emerged as the best-performing open-weight models in this task, with the 1.5 series outperforming the newer 2.0 and 2.5 versions on OCT, SLO, and CFP. Notably, the InternVL family demonstrated superior resistance to hallucinations compared to other model families, with HR scores exceeding 0.9 and showing the lowest hallucination rates across architectures. InternVL 2.5 also showed notable improvements on LP images, likely benefiting from additional RGB image data covering related diseases incorporated during training.  

An interesting observation is that the InternVL MPO variants (post-trained with reasoning preference) did not outperform their non-MPO counterparts. This contrasts with findings reported in the general domain~\cite{wu2024deepseekvl2mixtureofexpertsvisionlanguagemodels}. One possible explanation is that reasoning preferences in medicine differ~\cite{goh2024large}, and in this specific case, domain knowledge for interpreting anatomical structures in ophthalmology is arguably more important than reasoning optimization.  

By contrast, the LLaVA, Qwen, Yi, and DeepSeek families performed poorly across all modalities. While LLaVA-Med showed some improvement on CFP images, it failed on other
modalities such as SLO, likely due to its BioMedCLIP encoder being pretrained on microscopy and X-ray images but lacking exposure to scanning laser ophthalmoscopy data. Med-Flamingo, in turn, was ineffective across all five ophthalmic imaging modalities.

\textbf{Modality-Specific Performance Analysis.} 
We further compared results by imaging modality. 
As shown in Figure~\ref{fig:radar_chart}, SS was the most challenging modality for current MLLMs, with even the best-performing GPT-4o achieving only an F1 score of 0.2864, while most other models scored below 0.1. This suboptimal performance could be attributed to specific
challenges of this modality, such as frequent motion blur from rapid instrument and eye movements, and surgical tools occluding anatomical structures~\cite{2024_cataract1k_scidata}.  

In addition, for OCT, only GPT-4o, InternVL 1.5-2B, and InternVL 1.5-4B achieved somewhat meaningful results (precision > 0.8, F1 > 0.5). This observation underscores the domain knowledge needed for OCT interpretation, as these cross-sectional retinal images demand understanding of complex layered anatomical structures that differ significantly from the natural images typically used to pretrain vision-language models.  

By contrast, CFP and LP images proved the most accessible to the tested MLLMs, with most models achieving satisfactory results. Finally, SLO demonstrated the highest performance variation even within the same modal family. For example, LLaVA Vicuna-7B achieved an F1 score of 0.52, while other LLaVA variants performed poorly; similarly, InternVL 1.5-2B reached 0.53 F1, whereas InternVL 2.0-2B achieved nearly zero performance. This suggests that SLO interpretation success may depend heavily on specific model architectural choices and training strategies.

\begin{figure}[h!]
    \centering
    \includegraphics[width=0.8\textwidth]{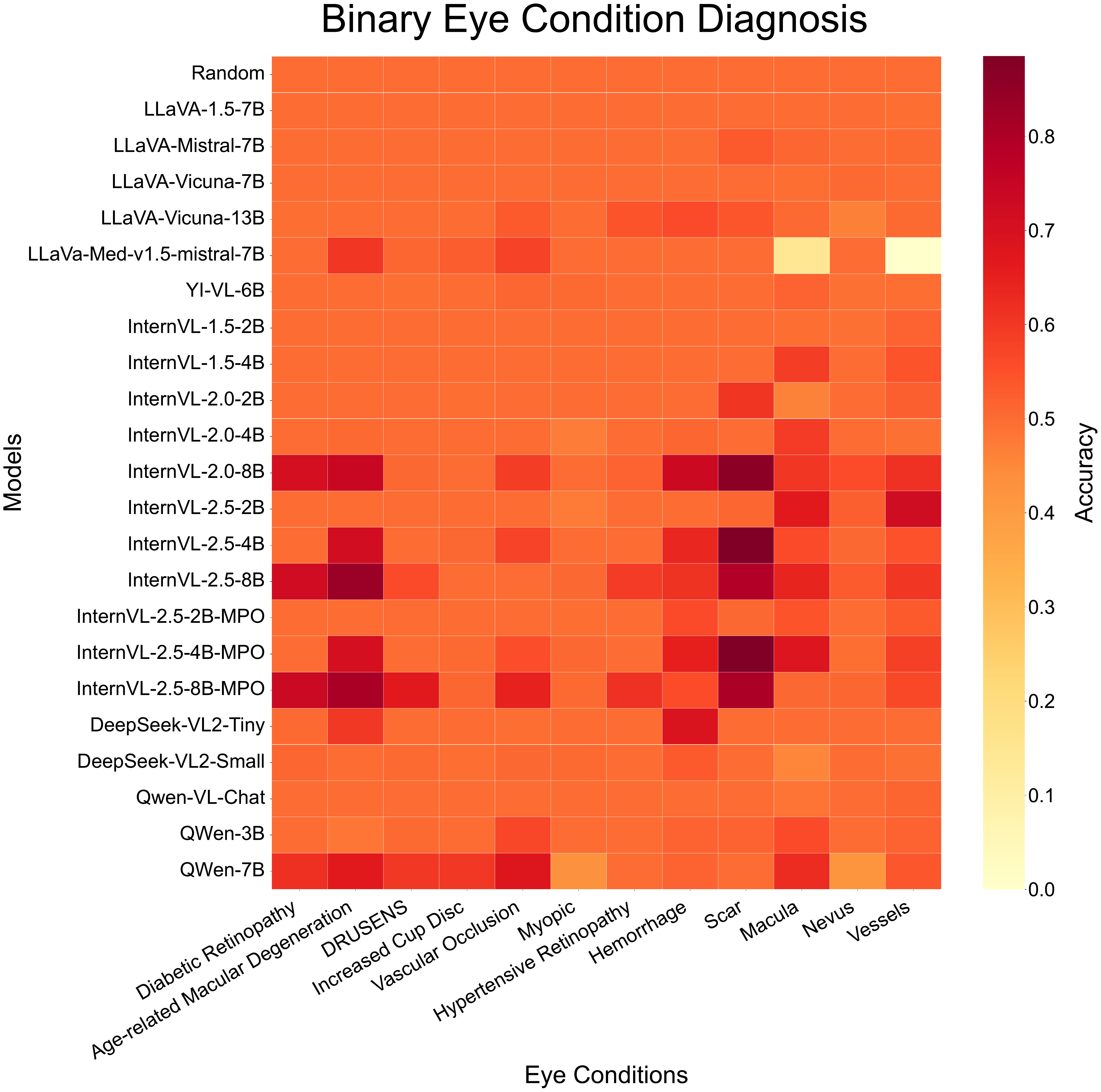}
    \caption{Binary Eye Condition Diagnosis Accuracy Heatmap. Performance comparison of 23 MLLMs across 12 eye conditions. Color scale represents classification accuracy (0-1), with darker colors indicating superior diagnostic performance.}
    \label{fig:binary_eye_condition_diagnosis_heatmap}
    \Description{Color-coded heatmap displaying binary classification accuracy of 23 MLLMs across 12 eye conditions, where darker cells indicate higher accuracy and lighter cells indicate lower accuracy.}
\end{figure}

\subsection*{Ophthalmic Disease Diagnosis}

\textbf{Overall Performance.} Table~\ref{tab:main_table_results} provides an overview of the ophthalmic disease diagnosis task, including both binary eye condition classification and multi-disease classification sub-tasks. As the datasets were selected with balanced distributions, we report accuracy as the primary metric, and Table~\ref{tab:main_table_results} also includes a random baseline for comparison.  

Several observations are noted. First, among open-weight general-domain models, Qwen-7B achieved the highest accuracy for binary eye condition diagnosis (58.26\%), followed by InternVL-2.5-8B (57.83\%). Moreover, InternVL-2.5-8B achieved the highest accuracy on multi-disease classification (35.95\%). Similar to the anatomical structure recognition task, medical MLLMs such as Med-Flamingo and LLaVA-Med did not outperform general models in either sub-task, with accuracies falling below the random baseline.  

Overall, the results indicate that both sub-tasks remain highly challenging for current models. For example, binary diagnosis performance across all MLLMs remained close to the random baseline, and a similar trend was observed for multi-disease classification.

\textbf{Binary Eye Condition Diagnosis.}  
Figure~\ref{fig:binary_eye_condition_diagnosis_heatmap} provides detailed comparisons of model performance across 12 eye conditions. InternVL-2.5-8B and InternVL-2.5-8B-MPO achieved above-chance accuracy (>50\%) on 11 and 12 conditions, respectively. For example, in detecting the presence of AMD, InternVL-2.5-8B achieved 83.61\% accuracy, while InternVL-2.5-8B-MPO reached 80.77\%. Likewise, Qwen-7B demonstrated relatively strong performance on specific conditions such as increased cup-to-disc ratio, vascular occlusion, and macular disorders.  

Despite these examples, overall performance across conditions remained suboptimal for all models. Critically, CFP was the primary imaging modality for this subtask. While the anatomical structure recognition results showed that models generally performed better on CFP compared to other modalities, their performance in disease diagnosis using CFP was considerably weaker. This suggests that reliable condition-specific diagnosis requires capabilities beyond structural recognition.

\begin{figure}[htbp]
\centering
\includegraphics[width=0.8\textwidth]{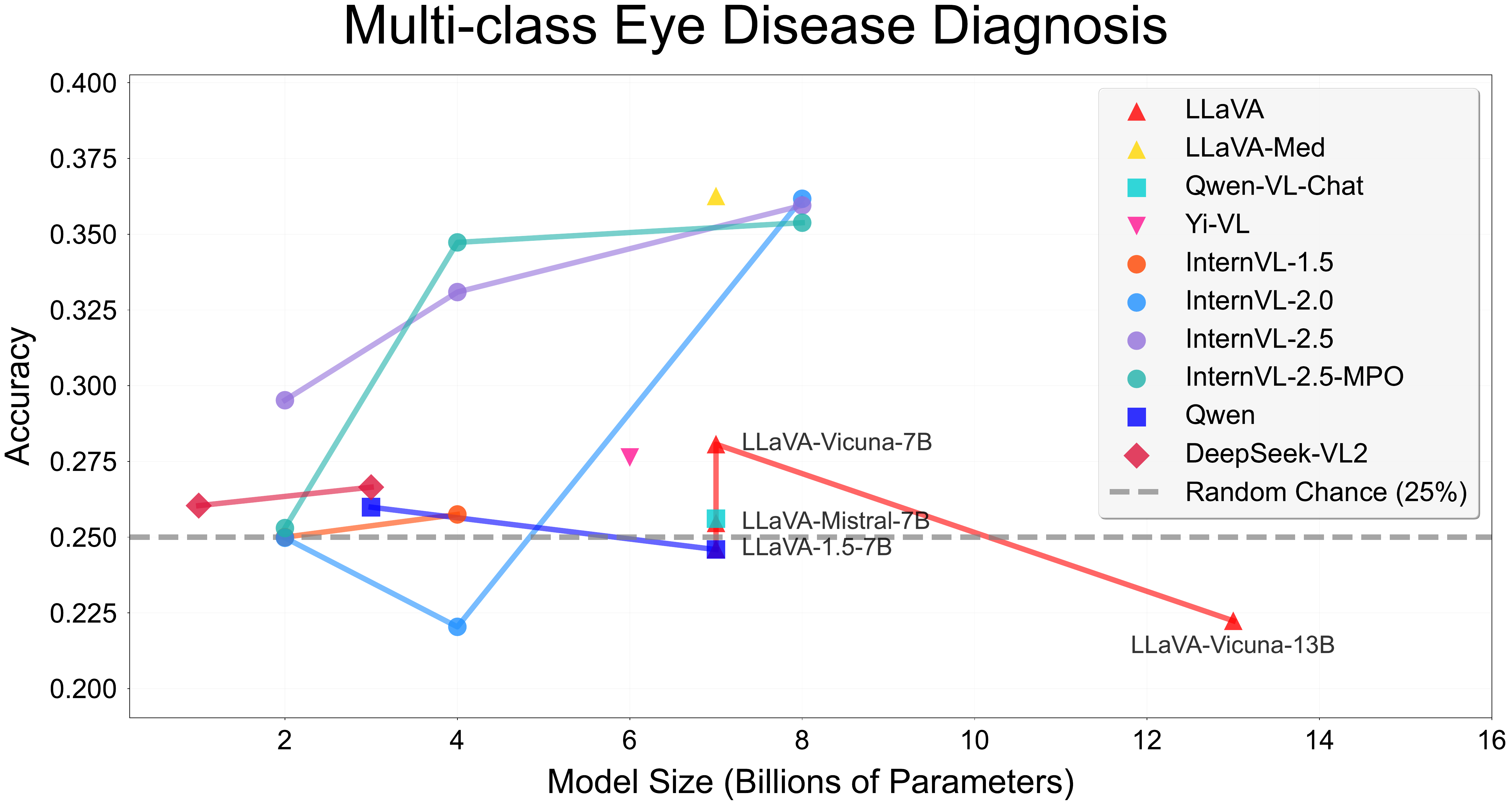}
\caption{Performance comparison of MLLMs on multi-class eye disease diagnosis task. The scatter plot shows the relationship between model size (billions of parameters) and diagnostic accuracy on a four-class eye disease classification task using CFP images. Each point represents a different model. Connected lines within each model family show the performance progression across different parameter scales. The gray dashed line indicates random chance performance (25\% for four-class classification). Selected LLaVA variants are labeled to distinguish between different architectural configurations.}
\label{fig:multiclass_diagnosis_scatter}
\Description{Scatter plot showing diagnostic accuracy versus model size in billions of parameters for multi-class eye disease classification, with connected lines within model families and a horizontal dashed line at 25 percent indicating random chance.}
\end{figure}

\textbf{Multi-Class Ophthalmic Disease Diagnosis.}  

Figure~\ref{fig:multiclass_diagnosis_scatter} presents the detailed results for the multi-disease classification sub-task, which aimed to distinguish between cataract, glaucoma, diabetic retinopathy, and normal conditions. As the results show, this task remained highly challenging for all models: the best-performing model achieved only 36.26\% accuracy, while most models scored near random chance levels (25\%).  

Figure~\ref{fig:multiclass_diagnosis_scatter} also compares model accuracy relative to model size and version. Within the InternVL family, performance improved progressively from the 1.5 series (25\% – 25.75\%) through the 2.0 series (22.04\% – 36.17\%) to the 2.5 series (29.52\% – 35.95\%). Consistent with the anatomical structure recognition task, the MPO variants underperformed relative to their standard counterparts.

\begin{figure}[htbp]
\centering
\includegraphics[width=0.8\textwidth]{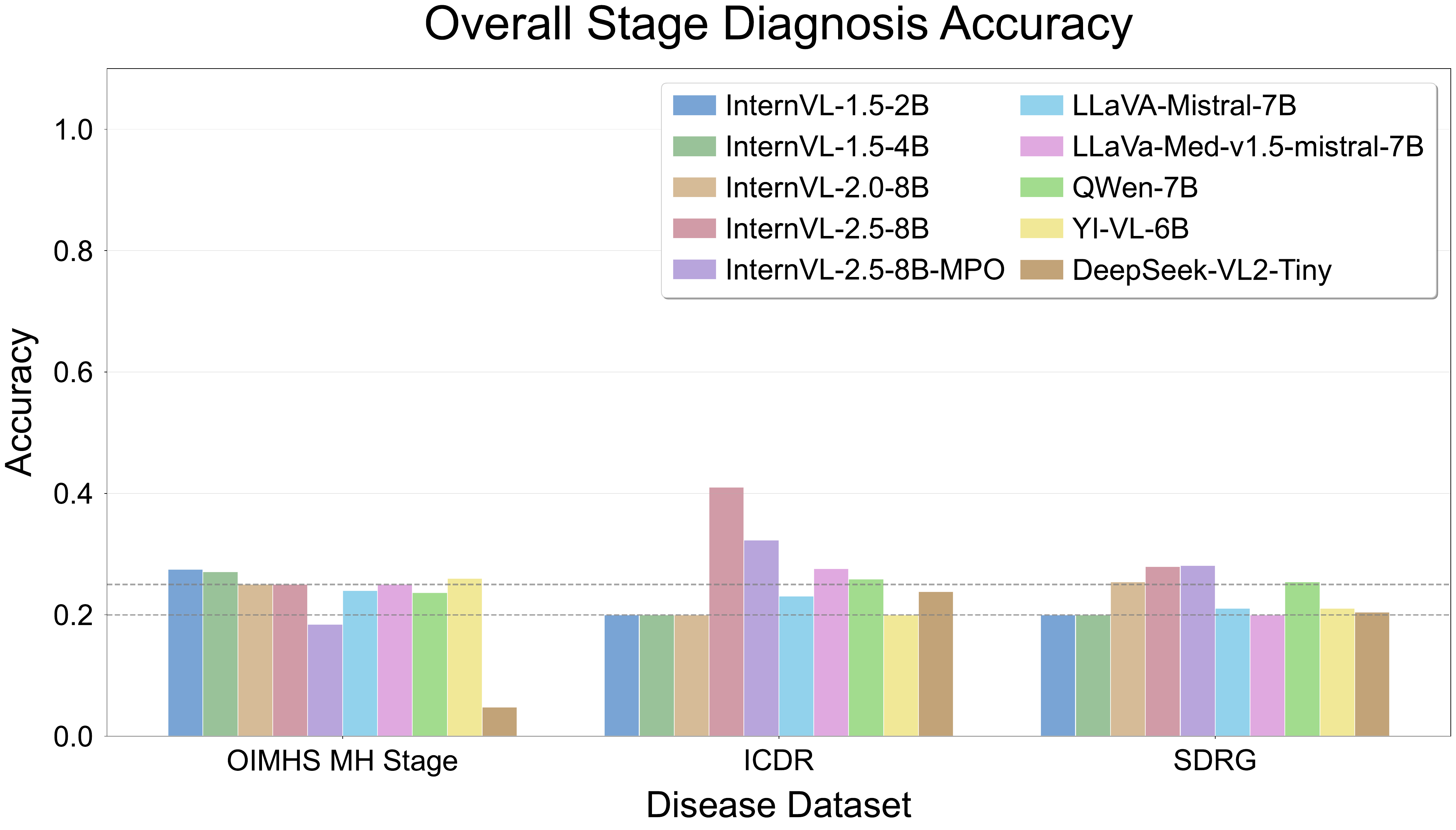}
\caption{Comparative Performance of MLLMs on Ophthalmic Stage Diagnosis Tasks. 
Bar chart comparing accuracy of 10 selected MLLMs across three distinct ophthalmic datasets requiring stage-based diagnosis: OIMHS Macular Hole (MH) Stage classification, ICDR severity grading, and SDRG. The horizontal dashed lines at 20\% and 25\% represent baseline performance thresholds. Models evaluated include InternVL variants (1.5-2B to 2.5-8B-MPO), LLaVA family models, LLaVA-Med-7B, Qwen-7B, Yi-VL-6B, and DeepSeek VL2-Tiny. ICDR demonstrates the highest achievable accuracies (up to 40\%), while OIMHS MH Stage and SDRG show more consistent performance in the 15\% - 25\% range. InternVL 2.5-8B exhibits superior performance on ICDR compared to other models.}
\label{fig:stage_diagnosis_accuracy}
\Description{Grouped bar chart comparing accuracy of 10 MLLMs on three stage diagnosis tasks (OIMHS MH Stage, ICDR, SDRG), with horizontal dashed lines at 20 and 25 percent baselines showing that most models perform near chance level.}
\end{figure}

\begin{figure*}[htbp]
\centering
\includegraphics[width=0.9\textwidth]{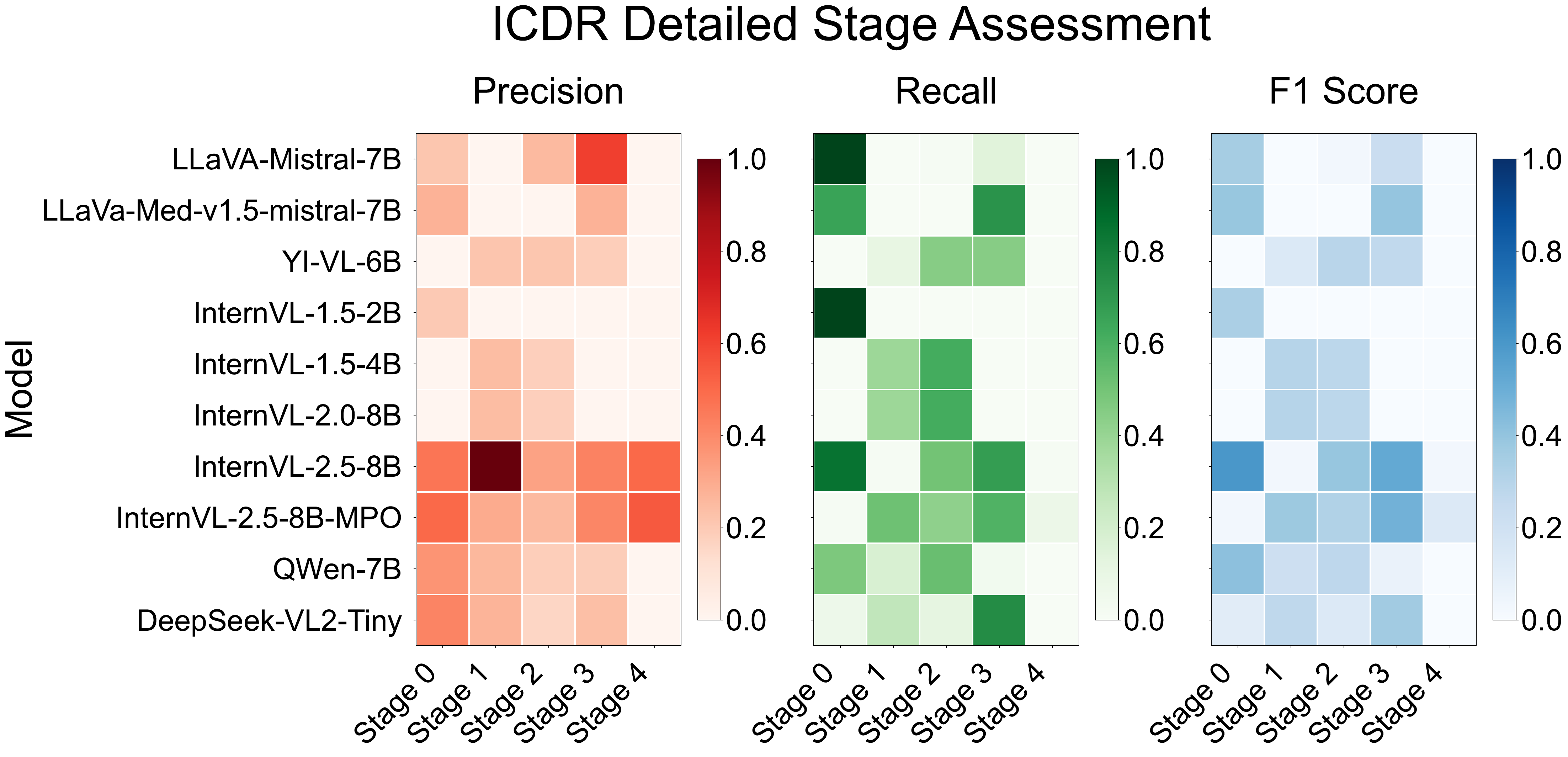}
\caption{\textbf{ICDR Detailed Stage Assessment Performance for MLLMs.} 
Heatmap visualization showing precision, recall, and F1 score of 10 selected MLLMs across five ICDR severity stages (0-4). Each subplot displays performance metrics as color-coded matrices, with darker colors indicating higher performance values (scale: 0.0-1.0). The precision matrix (left) shows models' ability to correctly identify specific stages, recall matrix (center) demonstrates sensitivity in detecting each stage, and F1 score matrix (right) provides balanced performance assessment. Overall performance varies significantly across stages, with Stage 0 and Stage 3 showing higher detectability compared to intermediate stages (1-2) across most models.}
\label{fig:stage_diagnosis_matrix}
\Description{Three side-by-side heatmaps showing precision, recall, and F1 scores of 10 MLLMs across five ICDR severity stages (0--4), with darker colors indicating higher values on a 0 to 1 scale.}
\end{figure*}

\begin{table*}[t]
\centering
\caption{Comprehensive results of ophthalmic disease staging assessment. Overall, model performance remained suboptimal across all stages.
}
\label{tab:stage_diagnosis_detailed}
\normalsize
\setlength{\tabcolsep}{3pt}
\renewcommand{\arraystretch}{0.8}
\resizebox{0.95\textwidth}{!}{
\newcommand{\hdrstrutA}{\rule{0pt}{2.8ex}}
\newcommand{\hdrstrutB}{\rule{0pt}{2.4ex}}
\begin{tabular}{l|l|c|c|c|c|c|c|c|c|c|c|c|c|c|c}
\toprule
\multirow{2}{*}{\textbf{Models}} & \multirow{2}{*}{\textbf{Metric}} & \multicolumn{4}{c|}{\textbf{OIMHS MH}} & \multicolumn{5}{c|}{\textbf{ICDR}} & \multicolumn{5}{c}{\textbf{SDRG}} \\
\cmidrule(lr){3-6} \cmidrule(lr){7-11} \cmidrule(lr){12-16}
&  & Stage 1 & Stage 2 & Stage 3 & Stage 4 & Stage 0 & Stage 1 & Stage 2 & Stage 3 & Stage 4 & Stage 0 & Stage 1 & Stage 2 & Stage 3 & Stage 4 \\
\midrule
\multirow{3}{*}{LLaVA-1.5-7B} & Precision & 0.0000 & 0.1571 & 0.2500 & \textbf{1.0000} & 0.0000 & 0.0000 & 0.0000 & 0.0000 & 0.2000 & 0.0000 & 0.0000 & 0.0000 & 0.1786 & 0.2371 \\
 & Recall & 0.0000 & \underline{0.9902} & 0.0029 & 0.0009 & 0.0000 & 0.0000 & 0.0000 & 0.0000 & \textbf{1.0000} & 0.0000 & 0.0000 & 0.0000 & \underline{0.4511} & \textbf{0.5865} \\
 & F1 & 0.0000 & 0.2712 & 0.0057 & 0.0018 & 0.0000 & 0.0000 & 0.0000 & 0.0000 & \textbf{0.3333} & 0.0000 & 0.0000 & 0.0000 & 0.2559 & \textbf{0.3377} \\
\midrule
\multirow{3}{*}{LLaVA-Mistral-7B} & Precision & 0.0048 & \underline{0.4000} & 0.1579 & \underline{1.0000} & 0.2120 & 0.0000 & 0.2500 & \underline{0.6111} & 0.0000 & 0.2069 & 0.0000 & 0.0000 & 0.2963 & 0.0000 \\
 & Recall & 0.9444 & 0.0073 & 0.0060 & 0.0024 & \textbf{1.0000} & 0.0000 & 0.0128 & 0.1410 & 0.0000 & 0.9925 & 0.0000 & 0.0000 & 0.0602 & 0.0000 \\
 & F1 & 0.0096 & 0.0143 & 0.0116 & 0.0049 & 0.3498 & 0.0000 & 0.0244 & 0.2292 & 0.0000 & 0.3424 & 0.0000 & 0.0000 & 0.1000 & 0.0000 \\
\midrule
\multirow{3}{*}{LLaVA-Vicuna-7B} & Precision & 0.0018 & 0.1896 & 0.1053 & 1.0000 & 0.2060 & 0.2222 & 0.0000 & 0.0000 & 0.1702 & 0.2025 & 0.1908 & 0.0000 & 0.0000 & 0.0000 \\
 & Recall & 0.3750 & 0.3564 & 0.0028 & 0.0014 & 0.7143 & 0.0769 & 0.0000 & 0.0000 & \underline{0.2051} & 0.5197 & \underline{0.4769} & 0.0000 & 0.0000 & 0.0000 \\
 & F1 & 0.0035 & 0.2476 & 0.0055 & 0.0028 & 0.3198 & 0.1143 & 0.0000 & 0.0000 & \underline{0.1860} & 0.2914 & \underline{0.2725} & 0.0000 & 0.0000 & 0.0000 \\
\midrule
\multirow{3}{*}{LLaVA-Vicuna-13B} & Precision & 0.0000 & 0.1424 & 0.1974 & 1.0000 & 0.2000 & 0.0000 & 0.0000 & 0.0000 & 0.0000 & 0.2031 & 0.1739 & 0.0000 & 0.0000 & 0.0000 \\
 & Recall & 0.0000 & 0.7574 & 0.1160 & 0.0005 & \underline{1.0000} & 0.0000 & 0.0000 & 0.0000 & 0.0000 & 0.9774 & 0.0301 & 0.0000 & 0.0000 & 0.0000 \\
 & F1 & 0.0000 & 0.2398 & 0.1461 & 0.0009 & 0.3333 & 0.0000 & 0.0000 & 0.0000 & 0.0000 & 0.3364 & 0.0513 & 0.0000 & 0.0000 & 0.0000 \\
\midrule
\multirow{3}{*}{LLaVA-Med-v1.5-mistral-7B} & Precision & \textbf{1.0000} & \textbf{1.0000} & 0.2703 & 1.0000 & 0.2772 & 0.0000 & 0.0000 & 0.2745 & 0.0000 & 0.2000 & 0.0000 & 0.0000 & 0.0000 & 0.0000 \\
 & Recall & 0.0000 & 0.0000 & \textbf{1.0000} & 0.0000 & 0.6538 & 0.0000 & 0.0000 & \underline{0.7179} & 0.0000 & \textbf{1.0000} & 0.0000 & 0.0000 & 0.0000 & 0.0000 \\
 & F1 & 0.0000 & 0.0000 & 0.4255 & 0.0000 & 0.3893 & 0.0000 & 0.0000 & 0.3972 & 0.0000 & 0.3333 & 0.0000 & 0.0000 & 0.0000 & 0.0000 \\
\midrule
\multirow{3}{*}{Qwen-VL-Chat} & Precision & 0.0000 & 0.0000 & 0.2698 & 0.6154 & 0.0000 & 0.1762 & 0.0000 & 0.0000 & 0.0000 & 0.0000 & 0.0000 & 0.2003 & 0.0000 & 0.0000 \\
 & Recall & 0.0000 & 0.0000 & 0.9933 & 0.0037 & 0.0000 & \textbf{0.8333} & 0.0000 & 0.0000 & 0.0000 & 0.0000 & 0.0000 & \textbf{1.0000} & 0.0000 & 0.0000 \\
 & F1 & 0.0000 & 0.0000 & 0.4244 & 0.0073 & 0.0000 & 0.2908 & 0.0000 & 0.0000 & 0.0000 & 0.0000 & 0.0000 & 0.3338 & 0.0000 & 0.0000 \\
\midrule
\multirow{3}{*}{Yi-VL-6B} & Precision & 0.0244 & 0.1544 & 0.2871 & 0.5722 & 0.0000 & 0.2162 & 0.2134 & 0.1872 & 0.0000 & 0.0000 & 0.1000 & 0.2011 & 0.2170 & 0.0000 \\
 & Recall & 0.0526 & 0.5033 & 0.1112 & \underline{0.3731} & 0.0000 & 0.1026 & 0.4487 & 0.4487 & 0.0000 & 0.0000 & 0.0075 & 0.2782 & \textbf{0.7669} & 0.0000 \\
 & F1 & 0.0333 & 0.2363 & 0.1603 & \underline{0.4517} & 0.0000 & 0.1391 & 0.2893 & 0.2642 & 0.0000 & 0.0000 & 0.0140 & 0.2334 & \textbf{0.3383} & 0.0000 \\
\midrule
\multirow{3}{*}{InternVL-1.5-2B} & Precision & 0.0075 & 0.1616 & 0.2984 & 1.0000 & 0.2000 & 0.0000 & 0.0000 & 0.0000 & 0.0000 & 0.2000 & 0.0000 & 0.0000 & 0.0000 & 0.0000 \\
 & Recall & 0.2105 & 0.7820 & 0.1064 & 0.0000 & 1.0000 & 0.0000 & 0.0000 & 0.0000 & 0.0000 & \underline{1.0000} & 0.0000 & 0.0000 & 0.0000 & 0.0000 \\
 & F1 & 0.0144 & 0.2679 & 0.1569 & 0.0000 & 0.3333 & 0.0000 & 0.0000 & 0.0000 & 0.0000 & 0.3333 & 0.0000 & 0.0000 & 0.0000 & 0.0000 \\
\midrule
\multirow{3}{*}{InternVL-1.5-4B} & Precision & 0.0052 & 0.1885 & \textbf{1.0000} & 1.0000 & 0.0000 & 0.2439 & 0.1798 & 0.0000 & 0.0000 & 0.0000 & 0.1667 & 0.2003 & 0.0000 & 0.0000 \\
 & Recall & 0.8421 & 0.2410 & 0.0000 & 0.0000 & 0.0000 & 0.3846 & 0.6154 & 0.0000 & 0.0000 & 0.0000 & 0.0075 & 0.9925 & 0.0000 & 0.0000 \\
 & F1 & 0.0103 & 0.2115 & 0.0000 & 0.0000 & 0.0000 & \underline{0.2985} & 0.2783 & 0.0000 & 0.0000 & 0.0000 & 0.0144 & 0.3333 & 0.0000 & 0.0000 \\
\midrule
\multirow{3}{*}{InternVL-2.0-2B} & Precision & 0.0000 & 0.1578 & 0.5000 & 1.0000 & 0.2042 & 0.0000 & 0.0000 & 0.0000 & 0.0000 & 0.2333 & 0.0982 & 0.0000 & 0.0000 & 0.0000 \\
 & Recall & 0.0000 & \textbf{0.9984} & 0.0010 & 0.0005 & 1.0000 & 0.0000 & 0.0000 & 0.0000 & 0.0000 & 0.9699 & 0.0827 & 0.0000 & 0.0000 & 0.0000 \\
 & F1 & 0.0000 & \underline{0.2725} & 0.0019 & 0.0009 & 0.3391 & 0.0000 & 0.0000 & 0.0000 & 0.0000 & 0.3761 & 0.0898 & 0.0000 & 0.0000 & 0.0000 \\
\midrule
\multirow{3}{*}{InternVL-2.0-4B} & Precision & 0.0000 & 0.1672 & 0.3068 & 0.0000 & 0.0000 & 0.0000 & 0.2000 & 0.0000 & 0.0000 & 0.0000 & 0.0000 & 0.2000 & 0.0000 & 0.0000 \\
 & Recall & 0.0000 & 0.8066 & 0.2694 & 0.0000 & 0.0000 & 0.0000 & \textbf{1.0000} & 0.0000 & 0.0000 & 0.0000 & 0.0000 & \underline{1.0000} & 0.0000 & 0.0000 \\
 & F1 & 0.0000 & \textbf{0.2769} & 0.2869 & 0.0000 & 0.0000 & 0.0000 & 0.3333 & 0.0000 & 0.0000 & 0.0000 & 0.0000 & 0.3333 & 0.0000 & 0.0000 \\
\midrule
\multirow{3}{*}{InternVL-2.0-8B} & Precision & 0.0000 & 0.1250 & 0.2709 & 0.0000 & 0.0000 & 0.2439 & 0.1798 & 0.0000 & 0.0000 & 0.2466 & 0.0000 & 0.0556 & 0.3106 & 0.0000 \\
 & Recall & 0.0000 & 0.0033 & 0.9981 & 0.0000 & 0.0000 & 0.3846 & 0.6154 & 0.0000 & 0.0000 & 0.9549 & 0.0000 & 0.0075 & 0.3083 & 0.0000 \\
 & F1 & 0.0000 & 0.0064 & \underline{0.4261} & 0.0000 & 0.0000 & 0.2985 & 0.2783 & 0.0000 & 0.0000 & \underline{0.3920} & 0.0000 & 0.0132 & \underline{0.3094} & 0.0000 \\
\midrule
\multirow{3}{*}{InternVL-2.5-2B} & Precision & 0.0008 & 0.0000 & 0.2717 & 1.0000 & 0.0000 & 0.0000 & 0.2104 & 0.1875 & \underline{0.5000} & 0.0000 & 0.0000 & 0.2043 & 0.0000 & \underline{0.7857} \\
 & Recall & 0.0526 & 0.0000 & 0.6836 & 0.0009 & 0.0000 & 0.0000 & 0.9872 & 0.0385 & 0.0513 & 0.0000 & 0.0000 & 1.0000 & 0.0000 & 0.0827 \\
 & F1 & 0.0016 & 0.0000 & 0.3889 & 0.0018 & 0.0000 & 0.0000 & 0.3468 & 0.0638 & 0.0930 & 0.0000 & 0.0000 & 0.3393 & 0.0000 & 0.1497 \\
\midrule
\multirow{3}{*}{InternVL-2.5-4B} & Precision & 0.0057 & 0.0000 & 0.0000 & 0.0000 & 0.2484 & 0.0000 & 0.1600 & 0.4737 & 0.0000 & 0.2226 & 0.0845 & 0.0000 & \textbf{1.0000} & 0.0000 \\
 & Recall & \textbf{1.0000} & 0.0000 & 0.0000 & 0.0000 & 1.0000 & 0.0000 & 0.0513 & 0.1154 & 0.0000 & 0.9624 & 0.0451 & 0.0000 & 0.0226 & 0.0000 \\
 & F1 & 0.0114 & 0.0000 & 0.0000 & 0.0000 & 0.3980 & 0.0000 & 0.0777 & 0.1856 & 0.0000 & 0.3616 & 0.0588 & 0.0000 & 0.0441 & 0.0000 \\
\midrule
\multirow{3}{*}{InternVL-2.5-8B} & Precision & 0.0000 & 0.0000 & 0.2709 & 0.0000 & \underline{0.4648} & \textbf{1.0000} & \textbf{0.3250} & 0.4240 & 0.5000 & 0.3213 & 0.0625 & 0.2016 & 0.3107 & \textbf{1.0000} \\
 & Recall & 0.0000 & 0.0000 & \underline{1.0000} & 0.0000 & 0.8462 & 0.0128 & 0.5000 & 0.6795 & 0.0128 & 0.9398 & 0.0226 & 0.1880 & 0.2406 & 0.0075 \\
 & F1 & 0.0000 & 0.0000 & \textbf{0.4263} & 0.0000 & \textbf{0.6000} & 0.0253 & \textbf{0.3939} & \textbf{0.5222} & 0.0250 & \textbf{0.4789} & 0.0331 & 0.1946 & 0.2712 & 0.0149 \\
\midrule
\multirow{3}{*}{InternVL-2.5-2B-MPO} & Precision & 0.0000 & 0.0000 & 0.0989 & 0.5965 & 0.0000 & 0.0000 & 0.2010 & \textbf{1.0000} & 0.0000 & 0.0000 & 0.0000 & 0.2098 & 0.0000 & 0.7419 \\
 & Recall & 0.0000 & 0.0000 & 0.1500 & \textbf{0.4879} & 0.0000 & 0.0000 & \underline{1.0000} & 0.0128 & 0.0000 & 0.0000 & 0.0000 & 1.0000 & 0.0000 & \underline{0.1729} \\
 & F1 & 0.0000 & 0.0000 & 0.1192 & \textbf{0.5367} & 0.0000 & 0.0000 & 0.3348 & 0.0253 & 0.0000 & 0.0000 & 0.0000 & \underline{0.3468} & 0.0000 & \underline{0.2805} \\
\midrule
\multirow{3}{*}{InternVL-2.5-4B-MPO} & Precision & 0.0058 & 0.0000 & 0.0000 & 0.0000 & 0.2686 & 0.0698 & 0.0750 & 0.4583 & 0.0000 & 0.2391 & 0.0792 & 0.0588 & \underline{0.6667} & 0.0000 \\
 & Recall & \underline{1.0000} & 0.0000 & 0.0000 & 0.0000 & 0.9744 & 0.0385 & 0.0385 & 0.1410 & 0.0000 & 0.9474 & 0.0602 & 0.0150 & 0.0150 & 0.0000 \\
 & F1 & 0.0116 & 0.0000 & 0.0000 & 0.0000 & \underline{0.4211} & 0.0496 & 0.0508 & 0.2157 & 0.0000 & 0.3818 & 0.0684 & 0.0240 & 0.0294 & 0.0000 \\
\midrule
\multirow{3}{*}{InternVL-2.5-8B-MPO} & Precision & 0.0000 & 0.0000 & 0.2615 & 0.6053 & \textbf{0.5000} & 0.2985 & \underline{0.2519} & 0.4107 & \textbf{0.5455} & \textbf{0.6250} & \underline{0.2804} & \underline{0.2386} & 0.2913 & 0.7500 \\
 & Recall & 0.0000 & 0.0000 & 0.7152 & 0.0105 & 0.0128 & \underline{0.5128} & 0.4231 & 0.5897 & 0.0769 & 0.0752 & \textbf{0.6767} & 0.3534 & 0.2782 & 0.0226 \\
 & F1 & 0.0000 & 0.0000 & 0.3830 & 0.0207 & 0.0250 & \textbf{0.3774} & 0.3158 & \underline{0.4842} & 0.1348 & 0.1342 & \textbf{0.3965} & 0.2848 & 0.2846 & 0.0438 \\
\midrule
\multirow{3}{*}{Qwen-3B} & Precision & 0.2394 & 0.2500 & \underline{1.0000} & 0.0000 & 0.2857 & \underline{0.5000} & 0.2012 & 0.3684 & 0.0000 & 0.0000 & \textbf{0.5833} & 0.2112 & 0.1786 & 0.0000 \\
 & Recall & 0.8947 & 0.0526 & 0.0526 & 0.0000 & 0.1538 & 0.0385 & 0.8333 & 0.0897 & 0.0000 & 0.0000 & 0.0526 & 0.9925 & 0.0376 & 0.0000 \\
 & F1 & \underline{0.3778} & 0.0870 & 0.1000 & 0.0000 & 0.2000 & 0.0714 & 0.3242 & 0.1443 & 0.0000 & 0.0000 & 0.0966 & \textbf{0.3483} & 0.0621 & 0.0000 \\
\midrule
\multirow{3}{*}{Qwen-7B} & Precision & \underline{0.2466} & 0.3333 & 0.0000 & 0.0000 & 0.3700 & 0.2593 & 0.1864 & 0.1875 & 0.0000 & \underline{0.3232} & 0.2396 & 0.1845 & 0.2286 & 0.0000 \\
 & Recall & 0.9474 & 0.0526 & 0.0000 & 0.0000 & 0.4744 & 0.1795 & 0.5256 & 0.0385 & 0.0000 & 0.4812 & 0.1729 & 0.4662 & 0.0602 & 0.0000 \\
 & F1 & \textbf{0.3913} & 0.0909 & 0.0000 & 0.0000 & 0.4157 & 0.2121 & 0.2752 & 0.0638 & 0.0000 & 0.3867 & 0.2009 & 0.2644 & 0.0952 & 0.0000 \\
\midrule
\multirow{3}{*}{DeepSeek-VL2-Tiny} & Precision & 0.0049 & 0.2098 & 0.2143 & 0.0000 & 0.4167 & 0.2727 & 0.1579 & 0.2377 & 0.0000 & 0.0000 & 0.0000 & 0.2073 & 0.1965 & 0.0000 \\
 & Recall & 0.7895 & 0.2672 & 0.0029 & 0.0000 & 0.0641 & 0.2692 & 0.1154 & \textbf{0.7436} & 0.0000 & 0.0000 & 0.0000 & 0.7669 & 0.2556 & 0.0000 \\
 & F1 & 0.0097 & 0.2350 & 0.0057 & 0.0000 & 0.1111 & 0.2710 & 0.1333 & 0.3602 & 0.0000 & 0.0000 & 0.0000 & 0.3264 & 0.2222 & 0.0000 \\
\midrule
\multirow{3}{*}{DeepSeek-VL2-Small} & Precision & -- & -- & -- & -- & 0.0000 & 0.0000 & 0.2150 & 0.0000 & 0.0000 & 0.1512 & 0.2394 & \textbf{0.2500} & 0.0000 & 0.5000 \\
 & Recall & -- & -- & -- & -- & 0.0000 & 0.0000 & 1.0000 & 0.0000 & 0.0000 & 0.6047 & 0.3148 & 0.0351 & 0.0000 & 0.0244 \\
 & F1 & -- & -- & -- & -- & 0.0000 & 0.0000 & \underline{0.3539} & 0.0000 & 0.0000 & 0.2419 & 0.2720 & 0.0615 & 0.0000 & 0.0465 \\
\bottomrule
\addlinespace[10pt]
\multicolumn{16}{l}{\normalsize \textit{Note:} \textbf{Bold} indicates the best performance in each column; \underline{underline} indicates the second best; ``--'' denotes inapplicable results} \\
\end{tabular}
}
\end{table*}

\subsection*{Ophthalmic Disease Staging Assessment}

\textbf{Overall Performance.} As shown in Table~\ref{tab:main_table_results}, 
the models overall demonstrated suboptimal performance on the disease staging task. 
The best-performing model, InternVL-2.5-8B, achieved an accuracy of only 26.67\%, which was only marginally above the random baseline. 
These results indicate that disease staging (classifying the severity level of a disease) 
is more challenging than simply detecting the presence or absence of a condition, 
which is consistent with clinical practice~\cite{ferris2013clinical}.

\textbf{Subset Analysis.} 
Figure~\ref{fig:stage_diagnosis_accuracy} shows the detailed staging accuracy of the top 10 models 
across the three subsets of this task: OIMHS, ICDR, and SDRG. 
The InternVL series consistently achieved the best performance across all three subsets; 
however, overall performance for all models remained suboptimal across the board.  

\textbf{Disease Staging Analysis.}  
Figure~\ref{fig:stage_diagnosis_matrix} and Table~\ref{tab:stage_diagnosis_detailed} provides stage-specific results for each disease. 
Overall, the models achieved higher F1-scores in the normal stage compared to other severity levels. 
For instance, InternVL-2.5-8B achieved Stage 0 F1 score of 0.6000 (ICDR) and 0.4789 (SDRG), 
substantially higher than for more severe stages. 
In contrast, other model families generally performed worse, with many models scoring at or below random baselines.

Collectively, suboptimal performance was consistently observed in both disease diagnosis and staging tasks. This underscores the substantial difficulty of distinguishing between multiple ocular pathologies and suggests that current models are not yet suitable for reliable application in eye disease screening or progression prediction without domain-specific training.

\subsection*{Demographic Information Inference}
\begin{table*}[t]
\centering
\caption{Patient demographic prediction from ophthalmic imaging. Across models, performance was near chance for both sex ($\approx$50\%) and age ($\approx$25\%), indicating no detectable demographic bias of MLLMs on ophthalmic imaging.}
\label{tab:sex_age_prediction}
\normalsize
\setlength{\tabcolsep}{3pt}
\renewcommand{\arraystretch}{1.2}

\resizebox{0.8\textwidth}{!}{
\begin{tabular}{l|c|c}
\toprule
\textbf{Models} & \textbf{Sex Accuracy} & \textbf{Age Accuracy} \\
\midrule
Random & 0.5000 & 0.2500 \\
GPT-4o & -- & -- \\
LLaVA-1.5-7B & \textbf{0.5105} & 0.2463 \\
LLaVA-Mistral-7B & 0.5000 & 0.2778 \\
LLaVA-Vicuna-7B & 0.5000 & 0.2350 \\
LLaVA-Vicuna-13B & 0.4993 & 0.1883 \\
LLaVA-Med-v1.5-mistral-7B & 0.5000 & 0.2500 \\
Qwen-VL-Chat & 0.5000 & \textbf{0.3457} \\
Yi-VL-6B & 0.5000 & 0.2538 \\
Med-Flamingo & -- & -- \\
InternVL-1.5-2B & 0.5000 & 0.2555 \\
InternVL-1.5-4B & 0.5000 & 0.2500 \\
InternVL-2.0-2B & 0.5000 & 0.2511 \\
InternVL-2.0-4B & 0.5059 & 0.2500 \\
InternVL-2.0-8B & 0.5004 & 0.2509 \\
InternVL-2.5-2B & 0.5000 & \underline{0.3076} \\
InternVL-2.5-4B & 0 & 0.2561 \\
InternVL-2.5-8B & \underline{0.5069} & 0.2367 \\
InternVL-2.5-2B-MPO & 0.5000 & 0.2661 \\
InternVL-2.5-4B-MPO & 0 & 0.2519 \\
InternVL-2.5-8B-MPO & 0 & 0.2965 \\
Qwen-3B & 0.5014 & 0.2500 \\
Qwen-7B & 0.4999 & 0.2517 \\
DeepSeek-VL2-Tiny & 0.4975 & 0.2500 \\
DeepSeek-VL2-Small & -- & 0.2050 \\
\bottomrule
\addlinespace[8pt]
\multicolumn{3}{l}{\normalsize \textit{Note:} \textbf{Bold} indicates the best performance in each column; \underline{underline} indicates the second best; ``--'' denotes inapplicable results.} \\
\end{tabular}
}
\end{table*}
As mentioned earlier, this task was included to evaluate potential model bias, specifically whether MLLMs could infer demographic information from ophthalmic imaging and use it
as the only information in decision-making. 
Table~\ref{tab:sex_age_prediction} presents the detailed performance of gender and age prediction. 
For gender prediction, all models demonstrated near-random performance, with accuracies hovering around the 50\% baseline. 
Similarly, for age prediction, performance remained close to random, indicating that the models were unable to extract demographic characteristics effectively from ophthalmic imaging. 
This finding contrasts with earlier work~\cite{poplin2018prediction,korot2021predicting} using CNN models. 
For instance, prior studies reported that CNNs fine-tuned on CFP data could predict sex with AUCs of 0.89–0.91 across different ethnic groups~\cite{betzler2021gender}, with anatomical features such as the foveal contour, optic nerve, and vascular arcades serving as discriminative markers~\cite{chueh2020prediction}. 
The key distinction lies in the training paradigm: CNNs were evaluated under supervised fine-tuning, whereas MLLMs were tested under a zero-shot setting as generative models. 
Future work should investigate the performance of MLLMs under supervised fine-tuning for demographic inference tasks and carefully examine potential biases that may arise in clinical applications.

\subsection*{Stability Analysis and Alternative Prompting Strategies}

To assess the reproducibility and robustness of our evaluation, we conducted additional experiments examining performance stability across random seeds and exploring alternative prompting strategies beyond the standard zero-shot approach.

\subsubsection*{Seed Stability Analysis}

We evaluated the anatomical recognition task using the OIMHS dataset with InternVL-1.5-2B as our testbed model. The original inference settings for InternVL did not contain randomness, using greedy decoding by default. To introduce randomness in model generation, we enabled stochastic sampling with the following configuration: \texttt{do\_sample: true}, \texttt{temperature: 0.7}, \texttt{top\_p: 0.9}, \texttt{num\_beams: 1}, and \texttt{max\_new\_tokens: 2048}. We then evaluated the model across three different random seeds (123, 124, and 125) to quantify the variance in performance metrics.

The results demonstrate considerable variation across seed settings, but consistently low overall performance. On the OIMHS anatomical recognition task, we observed the following macro-averaged metrics:
\begin{itemize}
    \item \textbf{Seed 123}: Precision = 0.118, Recall = 0.070, F1 = 0.082, HR = 0.974
    \item \textbf{Seed 124}: Precision = 0.112, Recall = 0.066, F1 = 0.074, HR = 0.926
    \item \textbf{Seed 125}: Precision = 0.224, Recall = 0.135, F1 = 0.160, HR = 0.977
\end{itemize}

While Seed 125 shows notably higher performance metrics, all three seeds yield F1 scores well below 0.2, indicating that anatomical recognition remains challenging for current MLLMs. The consistently high hallucination resistance (HR > 0.92) across all three seeds indicates that models reliably produce valid responses in the expected format, even when stochastic sampling is enabled. The variation in precision and recall reflects the inherent stochasticity of generative models under sampling-based decoding.

\subsubsection*{Alternative Prompting Strategies}

To explore whether performance could be improved through more sophisticated prompting techniques, we extended our evaluation to include three additional prompting strategies: few-shot prompting (3-shot), chain-of-thought (CoT) prompting, and self-consistency prompting. These experiments were conducted using InternVL-1.5-2B on the OIMHS anatomical recognition task.\newline
For each prompting strategy, we carefully constructed prompts that maintain the core task instruction while incorporating the specific reasoning or demonstration elements:\newline
\textbf{Chain-of-thought (CoT) prompting:} We prepended explicit step-by-step reasoning instructions to the base prompt: \textit{``Please analyze this image step by step, explaining your reasoning for each region identification. Think through the visual characteristics that help you identify each anatomical structure.''} This was followed by the standard task instruction: \textit{``This is an ophthalmology image of type OCT. Please identify and describe the type of each labeled bounding box in this image. Options can be Choroid, IRC, MH, Retina. Please just follow the format: Region ID: xxx; Type: xxx.''}\newline
\textbf{Few-shot prompting (3-shot):} We provided three demonstration examples before the test query. Each example showed the correct format and anatomical labels for a sample image: \textit{``Example 1: The correct answer for image 1 is: Region ID: 1; Type: Choroid; Region ID: 2; Type: Retina; Region ID: 3; Type: MH. Example 2: The correct answer for image 2 is: Region ID: 1; Type: Choroid; Region ID: 2; Type: Retina. Example 3: The correct answer for image 3 is: Region ID: 1; Type: Choroid; Region ID: 2; Type: Retina.''} This was followed by the task instruction for the test image.\newline
\textbf{Self-consistency prompting:} We prompted the model to analyze the image from multiple perspectives: \textit{``Please analyze this image using multiple approaches and provide your reasoning for each: Approach 1 - Anatomical Structure Analysis: Focus on the layered structure and identify regions based on anatomical knowledge. Approach 2 - Visual Pattern Recognition: Identify regions based on visual patterns, textures, and contrast differences. Approach 3 - Clinical Context Analysis: Consider the clinical significance and typical presentation of each structure in OCT. After analyzing with all three approaches, provide your final consolidated answer.''} This was followed by the standard task instruction.

The results reveal trade-offs and limitations across different prompting regimes:\newline
\textbf{Chain-of-thought (CoT) prompting} showed improved recall compared to the zero-shot baseline: Precision = 0.027, Recall = 0.237, F1 = 0.041, and HR = 0.018. The increased recall suggests that explicit reasoning instructions help the model identify more anatomical regions. However, the extremely low hallucination resistance reveals that CoT reasoning produces overly verbose outputs that frequently fail to follow the required structured format. While CoT appears to improve the model's region detection capability, the poor format compliance makes it impractical without additional output parsing.\newline
\textbf{Few-shot prompting (3-shot)} yielded: Precision = 0.230, Recall = 0.269, F1 = 0.235, and HR = 0.741. Compared to the zero-shot baseline (Precision = 0.109, Recall = 0.109, F1 = 0.109, HR = 1.0), few-shot examples improved both precision and recall. However, the decreased hallucination resistance (from 1.0 to 0.741) indicates that longer prompts with multiple examples occasionally lead to format violations or incomplete responses, suggesting a trade-off between improved performance and output reliability.\newline
\textbf{Self-consistency prompting} produced: Precision = 0.027, Recall = 0.056, F1 = 0.027, and HR = 0.016. Similar to CoT, the very low hallucination resistance indicates that prompting for multiple reasoning approaches generates outputs that severely violate format constraints. The multi-perspective reasoning instructions appear to confuse the model, resulting in both poor task performance and unreliable output structure.
Overall, while seed variations show performance fluctuations within a consistently low range, chain-of-thought prompting shows some promise for improving recall in anatomical recognition tasks. However, few-shot prompting decreased overall performance when accounting for format compliance, and self-consistency prompting degraded both accuracy and output reliability. These results suggest that current ophthalmology MLLMs require better mechanisms to balance complex reasoning instructions with structured output generation before advanced prompting strategies can be reliably deployed in clinical settings.

\section{Discussion}

First, MLLMs—including both general-domain and medical-domain models—demonstrated suboptimal performance across ophthalmic tasks, with average scores of 0.2113 F1 for anatomical recognition, 51.86\% accuracy for binary disease diagnosis, 28.23\% accuracy for multi-class disease diagnosis, and 21.05\% accuracy for disease staging. Many of these results were only marginally above random baselines. The overall performance was substantially lower than what has been reported in other domains~\cite{zhang2024mm,yin2024survey}.
To further validate these findings, we selected subsets of \dataset and reformulated the tasks as classification problems to train CNN models. For anatomical recognition, we cropped individual anatomical regions and assigned corresponding labels; for disease diagnosis, we focused on glaucoma detection; and for staging assessment, we selected macular hole (MH) staging. We then fine-tuned a CNN model for each task and the accuracies are presented in Table~\ref{tab:cnn_results}. The CNNs achieved consistently high performance, with accuracies ranging from 80\% to 98\%. These results align with previous literature~\cite{chen2019multi,gao2024recent} and further demonstrate that \dataset is clearly learnable.
Collectively, these findings highlight the significant challenges of applying current MLLMs to ophthalmology in zero-shot settings and suggest that domain-specific training may be necessary to achieve clinically meaningful performance.

\begin{table*}[tbp!]
\centering
\caption{Performance of supervised CNNs on anatomical recognition across multiple imaging modalities and selected diagnostic tasks. The high accuracies support the separability of the \dataset dataset.}
\resizebox{\textwidth}{!}{
    \begin{tabular}{lcccccccc}
    \toprule
    & \multicolumn{6}{c}{\textbf{Anatomical Recognition}} & \textbf{Disease Diagnosis} & \textbf{Staging Assessment}\\
    \cmidrule(lr){2-7} \cmidrule(lr){8-8} \cmidrule(lr){9-9}
    \textbf{Model} & \textbf{Macro Avg} & \textbf{SS} & \textbf{OCT} & \textbf{SLO} & \textbf{LP} & \textbf{CFP} & \textbf{Glaucoma} & \textbf{MH Stage} \\
    \midrule
    CNN & 0.9436 & 0.9376 & 0.9842 & 0.9492 & 0.8885 & 0.9586 & 0.8269 & 0.9817 \\
    \bottomrule
    \end{tabular}
}
\label{tab:cnn_results}
\end{table*}

In addition, the results consistently show that medical MLLMs may not perform well in specific medical specialties. For instance, Table~\ref{tab:main_table_results} shows that LLaVA-Med achieved suboptimal performance on anatomical recognition and binary diagnosis, underperforming the general LLaVA variants in direct comparisons. Prior studies on language-only tasks in ophthalmology (e.g., text summarization and knowledge testing) have also reported similar findings—that domain-specific medical LLMs do not necessarily outperform general-domain models in this field~\cite{gilson2024language}. Specifically, as the results demonstrate, LLaVA-Med could identify OCT and CFP image types but failed at higher-level tasks such as anatomical recognition and disease diagnosis.
One possible explanation is that LLaVA-Med was adapted using images and text from PubMed literature. While this may provide models with basic knowledge of imaging modalities, it is insufficient for learning the deeper structural and disease-specific features required for ophthalmology. To further examine this, we fine-tuned LLaVA-Med on a balanced subset of OCT and CFP images. Following its established training strategy~\cite{2024_neurips_llava,2024_neurips_llavamed}, we froze the visual encoder and fine-tuned the MLP adapter and LLM. However, fine-tuning led to poor results: the model produced repetitive outputs for anatomical recognition and empty responses for diagnostic tasks. This suggests that the limitations may also be related to architectural constraints and to its training strategies (LLaVA-Med relies on the original LLaVA architecture, which may be outdated).

\begin{table*}[tbp!]
\centering
\caption{Error taxonomy for MLLMs in ophthalmic diagnosis. Text generation failures, medical knowledge errors, and inconsistent reasoning primarily target textual generation errors in ophthalmic tasks, while misinterpreted visual features and absent visual processing focus on MLLMs' understanding of ophthalmic images and vision-language alignment errors.}

\label{tab:mllm_error_taxonomy}
\resizebox{\textwidth}{!}{
\begin{tabular}{lcccc}
\toprule
\\[0.5ex]
\textbf{\large Error Type} & \textbf{\large Definition} & \textbf{\large Identification Criteria} & \textbf{\large Typical Examples} & \textbf{\large Clinical Impact} \\[1.5ex]
\midrule
\textbf{1. Text Generation} & Complete breakdown of language & • Infinite token repetition & • "sign sign sign sign..." & \textbf{Critical} \\
\textbf{Failures} & model text generation mechanism, & • Corrupted output with & • "planations: planations:..." & Completely unusable, \\
& producing incomprehensible output & visible special tokens & • Visible artifacts & requires regeneration \\
& & • Complete text structure destruction & • "<end\_of\_sentence>" & \\[1.5ex]
\midrule
\textbf{2. Medical Knowledge} & Fundamental misunderstanding & • Disease feature confusion & • "GLAUCOMA: YES; shows & \textbf{High} \\
\textbf{Errors} & of medical facts, disease & • Medical terminology misuse & microaneurysms, hemorrhages" & Incorrect medical facts \\
& concepts, or anatomical knowledge & • Anatomical structure errors & (describing diabetic retinopathy & may lead to wrong \\
& & • Pathophysiology confusion & as glaucoma) & treatment decisions \\[1.5ex]
\midrule
\textbf{3. Inconsistent} & Logical contradictions in & • Prediction contradicts evidence & • "GLAUCOMA: NO; But shows & \textbf{High} \\
\textbf{Reasoning} & reasoning process despite & • Conflicting statements within & elevated IOP and optic cupping" & Logical contradictions \\
& correct medical knowledge & same response & • "GLAUCOMA: YES; Optic disc & may mislead clinical \\
& & • Inconsistent logical chain & appears completely normal" & decision-making \\[1.5ex]
\midrule
\textbf{4. Misinterpreted} & Model processes visual content & • Uses visual description vocabulary & \textbf{Visual Omission:} & \textbf{Medium-High} \\
\textbf{Visual Features} & and describes image features & • Mentions specific anatomy & • "Clear disc boundaries, normal & Shows vision-language \\
& but incorrectly interprets & • Describes visual attributes & cup-to-disc ratio" (GT: Glaucoma) & integration but incorrect \\
& clinical significance & • Spatial relationship description & \textbf{Visual Hallucination:} & clinical judgment \\
& & • Wrong clinical interpretation & • "Obvious optic cupping visible" & False negatives delay \\
& & & (GT: Non-Glaucoma) & diagnosis; false positives \\
& & & \textbf{Feature Misinterpretation:} & cause overtreatment \\
& & & • "Disc pale, but normal variation" & \\[1.5ex]
\midrule
\textbf{5. Absent Visual} & Model does not perform actual & • No specific visual description & \textbf{Medical Template:} & \textbf{Medium} \\
\textbf{Processing} & visual content analysis, & • Generic medical content & • "Glaucoma is a group of diseases & No diagnostic value but \\
& relying on generic knowledge & • Procedural language & that can cause..." (identical text) & typically does not \\
& or avoidance strategies & • Avoidance expressions & \textbf{Visual Avoidance:} & directly mislead \\
& & & • "Image not clear enough" & May cause delays \\
& & & \textbf{Procedural Deflection:} & in care-seeking \\
& & & • "Need comprehensive examination" & \\
\bottomrule
\end{tabular}
}
\end{table*}
\begin{figure*}[tp]
    \centering
    \includegraphics[width=\textwidth]{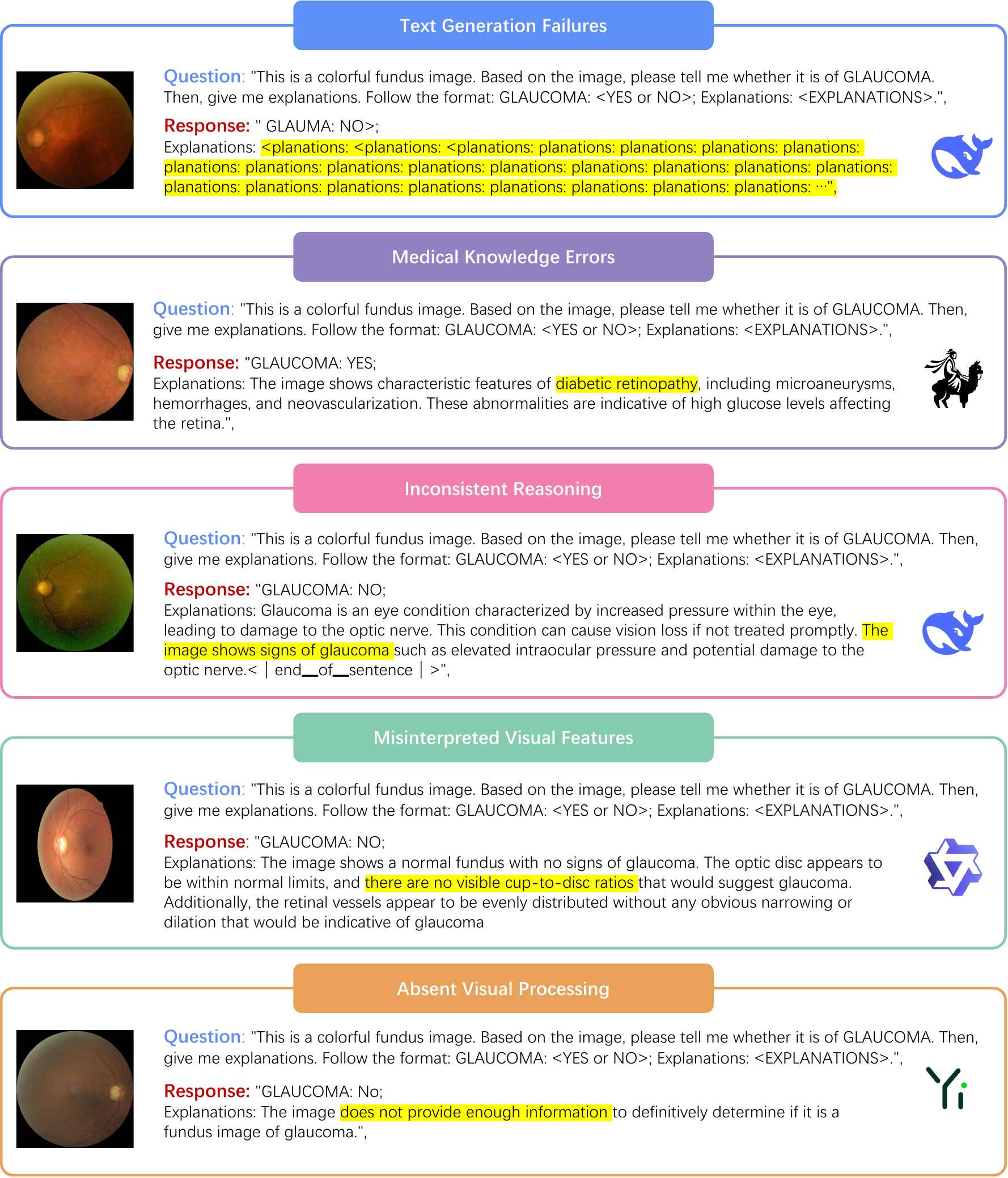}
    \caption{Five representative error cases. Each case illustrates a distinct error type: (a) Text generation failure case showing severe output collapse with infinite repetition of "planations"; (b) Medical knowledge error case where glaucoma is diagnosed but the explanation describes diabetic retinopathy signs; (c) Inconsistent reasoning case with contradictory final diagnosis ("NO") despite describing glaucomatous signs; (d) Misinterpreted visual features case failing to detect glaucomatous changes in a confirmed glaucoma patient; (e) Absent visual processing case where the system claims insufficient image information rather than performing analysis.}
    \label{fig:error_example}
    \Description{Five panels showing representative MLLM error cases in glaucoma diagnosis: (a) text generation collapse with repeated tokens, (b) wrong disease explanation despite correct label, (c) contradictory reasoning and diagnosis, (d) failure to detect glaucomatous features, and (e) refusal to analyze the image.}
\end{figure*}

\begin{table}
\centering
\caption{Distribution of error types on the 100 subset of glaucoma diagnosis. Misinterpreted Visual Features accounts for the highest proportion at 50\%, indicating that current MLLMs have insufficient capability in understanding ophthalmic images.}
\resizebox{0.5\textwidth}{!}{
\begin{tabular}{l r r}
\toprule
\textbf{Error Types} & \textbf{Counts} & \textbf{Proportion (\%)} \\
\midrule
Misinterpreted Visual Features & 50 & 50.0 \\
Inconsistent Reasoning & 21 & 21.0 \\
Absent Visual Processing & 15 & 15.0 \\
Text Generation Failures & 8 & 8.0 \\
Medical Knowledge Errors & 6 & 6.0 \\
\bottomrule
\end{tabular}
}
\label{tab:error_distribution}
\end{table}
We further conducted a detailed error analysis to systematically characterize the types of errors produced by MLLMs in ophthalmology. 
Specifically, we focused on glaucoma diagnosis and sampled 100 failure cases from the 11,301 errors produced by the 23 MLLMs that provided meaningful responses,\footnote{GPT-4o was not included in this analysis, as it did not participate in the diagnosis task.} using established diagnostic criteria~\cite{gulshan2016development,hendrycks2019using,devlin2019bert,maynez2020faithfulness,rudin2019stop,mckinney2020international}. Following these studies, we applied a combination of automatic and manual review, using GPT-4o as an evaluator and supplementing with manual verification.
Table~\ref{tab:mllm_error_taxonomy} summarizes the primary error categories, along with their definition, identification criteria, typical examples and clinical impact. Overall, we identified five major error types and the distribution is shown in Table~\ref{tab:error_distribution}, including text generation failures, medical knowledge errors, inconsistent reasoning, misinterpreted visual features and absent visual processing. 
Among these, the first three categories primarily target textual generation errors in ophthalmic tasks, while misinterpreted visual features and absent visual processing focus on MLLMs' understanding of ophthalmic images and vision-language alignment errors. 
Representative cases for each of the error types are illustrated in Figure~\ref{fig:error_example}.
Our analysis reveals several critical findings regarding MLLM performance in ophthalmic diagnosis. Misinterpreted visual features emerged as the most prevalent error type, accounting for 50\% of all failures, indicating that while models process visual content, they frequently arrive at incorrect diagnoses due to inadequate understanding of ophthalmic imagery. Inconsistent reasoning constituted 21\% of errors, manifesting as logical contradictions between final diagnoses and supporting explanations, or the generation of generic responses that lack clinical specificity. Absent visual processing failures occurred in 15\% of cases, where models failed to properly analyze fundus images despite claiming to do so. Text generation failures, comprising 8\% of errors, involved complete system breakdowns with infinite repetition patterns, predominantly observed in DeepSeek models. Medical knowledge errors represented 6\% of failures, reflecting basic terminology and domain knowledge deficiencies. 
Model-specific patterns emerged across different architectures: DeepSeek exhibited susceptibility to text generation collapse, InternVL demonstrated high rates of false positives and negatives in visual interpretation, LLaVA-Med showed systematic bias toward positive diagnoses regardless of image content, Yi-VL often provided superficial visual analysis, while Qwen, despite generally superior performance, still exhibited significant interpretation errors in complex ophthalmic cases.
These findings provide preliminary insights that can inform the design of domain-specific training strategies and facilitate more rigorous evaluation of MLLMs in ophthalmic applications.

Despite these challenges, it is also important to emphasize that the zero-shot setting is particularly difficult: a single model must handle diverse applications, heterogeneous input modalities, and varied output formats without any fine-tuning. This is considerably more demanding than the traditional paradigm, in which a separate model is fine-tuned for each specific task. Even under this setting, some promising observations are noted. For example, Qwen and InternVL demonstrated potential in eye disease screening across 12 eye conditions, achieving overall accuracies of 58.26\% and 57.83\%, respectively, with a single model. InternVL also showed encouraging performance in anatomical recognition. Notably, these models are relatively lightweight (2B–8B parameters), suggesting potential feasibility for local deployment in resource-constrained environments, while also mitigating privacy concerns compared with using proprietary API-based models.

To support further development and evaluation, we publicly release \dataset and its associated pipeline to the community. In addition to the full dataset of 32,633 instances, we provide a 1,000-instance representative subset, referred to as the LMOD+ subset, which covers all five imaging modalities and task categories to enable rapid evaluation. Users can employ this subset for quick benchmarking before scaling to the full dataset. We also release an evaluation leaderboard to facilitate transparent and efficient performance assessment.

\section{Conclusion}
In this study, we present \dataset, a comprehensive multimodal dataset with multi-granular annotations across 32,633 instances spanning five key imaging modalities, anatomical structures, free text, and demographic information, tailored for MLLMs and generative models.
We propose a unified and systematic data curation pipeline that repurposes datasets originally designed for earlier models and adapts them for MLLM development and evaluation.
\dataset covers 12 common ophthalmic conditions and supports key applications, including anatomical structure recognition, disease screening, disease staging, and demographic prediction for potential bias evaluation.
We systematically evaluated 24 state-of-the-art MLLMs to characterize both the potential and limitations of their adoption in ophthalmology.
Finally, we publicly release \dataset and the associated data pipeline to the community, enabling direct application to emerging datasets and models and supporting further development.

Our study has several primary limitations. 
First, while we systematically evaluated 24 models, the rapid pace of model development makes it impossible to cover every new release. 
To address this, we have made \dataset and its pipeline publicly available so that the community can readily apply them to emerging models.  
Second, although we included key ophthalmic applications ranging from anatomical structure recognition, disease screening, and disease staging to demographic prediction for potential bias evaluation, other tasks—such as treatment plan generation—are also important for comprehensive ophthalmic patient management~\cite{olawade2025enhancing}. 
A major challenge, however, is that most available ophthalmic datasets are primarily image-focused and lack patient information or clinical notes due to privacy constraints~\cite{khan2021global}. 
Developing multimodal datasets that facilitate AI-assisted end-to-end ophthalmic patient management will therefore be an important direction for future work.  
Finally, we encourage broader community efforts in the development and evaluation of MLLMs to advance ophthalmic applications and ultimately reduce the global burden of vision-threatening diseases with the assistance of AI.

\begin{acks}
This study was supported by grant R00LM014024 from the National Library of Medicine, National Institutes of Health; the Intramural Research Program of the National Library of Medicine and the National Eye Institute, National Institutes of Health; and a Google Research Scholar Award. 
\end{acks}

\bibliographystyle{ACM-Reference-Format}
\bibliography{refs_ACM}

\appendix

\section*{Appendix}

\section{Experimental Setup}
We developed a general framework based on PyTorch, providing a unified interface for performing inference across various MLLMs. This framework ensures consistent evaluation and smooth integration with different models.

For each MLLM, we used the same computing infrastructure—specifically, two RTX 6000 GPUs—to perform the inference. We evaluated the models using ten different ophthalmology datasets, with consistent prompts and inputs provided to each MLLM. Moreover, we applied the default hyperparameters for each model during the evaluation. This approach allowed us to fairly compare the performance of the different models.

\section{Computational Resource}
The computing infrastructure includes 11 GPU nodes, each equipped with 2x AMD EPYC 7742 processors (128 cores), 1TB of RAM, and 8 Quadro RTX 6000 GPUs per node. Additionally, there are 7 GPU nodes with 2x Intel Icelake Xeon Platinum 8358 processors.

For MLLMs inference tasks on various ophthalmology datasets, the runtime typically ranges from two to four hours, depending on the specific dataset.

\section{Use Of AI Assistants}
We used AI tools to assist with coding tasks, such as debugging and optimizing code during the development phase. Additionally, we leveraged AI to help polish the manuscript, addressing grammar issues and ensuring clarity and coherence in our presentation. However, all critical decisions such as the research design, methodology, and conclusions were made independently by the authors.

\section{Hyperparameters}

This section outlines the essential hyperparameters that were chosen for the MLLMs in our experiments. 

\begin{enumerate}

\item \textbf{Image Resolution}: The image resolution defines the size of the visual input processed by each MLLM. Higher resolutions capture finer details.
\item \textbf{Top-p Sampling}: Top-p, also known as nucleus sampling, is a hyperparameter that influences the randomness of a language model's output. It defines a probability threshold and selects the smallest set of tokens whose cumulative probability exceeds this threshold. The model then samples randomly from this subset to generate the output. This approach allows for more diverse and creative results compared to methods that randomly sample from the entire vocabulary.

\item \textbf{Temperature}: The temperature hyperparameter influences the randomness of the model's output by scaling logits before applying softmax. Higher temperatures (e.g., >1) encourage more diverse outputs by flattening the probability distribution, making it suitable for creative tasks. Lower temperatures (e.g., <1) concentrate the distribution, resulting in more focused outputs, which is critical in medical domains to ensure reliable, deterministic responses. Temperature is disabled when set to 0. 

\item \textbf{Beams Number}: Beam search is a decoding strategy that retains multiple candidate sequences at each generation step. A higher number of beams (e.g., 5 or 10) explores more possibilities, potentially yielding better results at the cost of increased computation. A lower beams number (e.g., 1) favors efficiency and speed but risks missing better sequences, which may be a concern in domains requiring high-quality outputs.

\item \textbf{Number of Parameters}: The number of parameters refers to the total count of learnable weights in a model, directly influencing its capacity and performance. Larger models tend to perform better due to increased capacity, but at the cost of higher memory usage and slower inference times.

\item \textbf{Max New Tokens}: This hyperparameter limits the number of tokens generated by the model during inference.

\end{enumerate}

\section{Supervised Training Settings}

To evaluate the feasibility of our proposed benchmark, we implemented neural network classifiers for anatomical recognition and diagnosis analysis. For both tasks, we used 80\%, 15\%, and 5\% data for training, validation, and test. For anatomical recognition, we employed a CNN visual encoder whose architecture is like below: 



\noindent The CNN was trained with the following settings:
\begin{itemize}
    \item Image resolution: 128 $\times$ 128
    \item Batch size: 512
    \item Learning rate: 0.001
    \item Epochs: 20
\end{itemize}

For diagnostic analysis, we fine-tuned RETFound as the visual encoder. RETFound is a foundation model for retinal images, built on a large Vision Transformer (ViT) architecture with 24 Transformer blocks and an embedding vector size of 1,024 \cite{2023_nature_eye_foundation}. The RETFound model offers two variations designed for different image types: CFP 
and OCT.
For macular hole (MH) stage classification, we employed the OCT variation, while the CFP model was used for glaucoma classification (according to the dataset's image type).
For both tasks, we fine-tuned RETFound using the default parameter settings:
\begin{itemize}
    \item Image resolution: 224 $\times$ 224
    \item Batch size: 16
    \item Base learning rate: 5e-3
    \item Epochs: 50
    \item Layer decay: 0.65 
    \item Weight decay: 0.05 
    
\end{itemize}

The model's performance on anatomical recognition and diagnosis analysis tasks served as a baseline for the complexity of our dataset, and is compared with the performance of MLLMs in subsequent sections.

\section{Details of Balanced Evaluation Dataset}

\begin{table}[ht]
\centering
\caption{Subsample sizes used for balanced evaluation. Counts are per class/stage/group.}
\label{tab:subsample_balanced_sizes}
\begin{tabular}{l l l c}
\toprule
\textbf{Parent dataset} & \textbf{Subspecialty} & \textbf{Label schema} & \textbf{Count (per class)} \\
\midrule
\multicolumn{4}{c}{\textbf{Disease diagnosis}}\\
\midrule
ORIGA         & Glaucoma                 & 2-class   & 168  \\
Eye Disease   & Multi-class               & 4-class   & 1007 \\
Eye Disease   & Cataract            & 2-class   & 1038 \\
Eye Disease   & Diabetic retinopathy                  & 2-class   & 1074 \\
Eye Disease   & Glaucoma            & 2-class   & 1007 \\
BRSET         & Diabetic retinopathy                  & 2-class   & 1070 \\
BRSET         & AMD                 & 2-class   & 299  \\
BRSET         & Drusens             & 2-class   & 2830 \\
BRSET         & Increased cup disc  & 2-class   & 3204 \\
BRSET         & Vascular occlusion  & 2-class   & 101  \\
BRSET         & Myopic              & 2-class   & 270  \\
BRSET         & Hypertensive retinopathy & 2-class & 283 \\
BRSET         & Hemorrhage          & 2-class   & 95   \\
BRSET         & Scar                & 2-class   & 291  \\
BRSET         & Macula              & 2-class   & 4675 \\
BRSET         & Nevus               & 2-class   & 130  \\
BRSET         & Vessels             & 2-class   & 806  \\
\addlinespace
\multicolumn{4}{c}{\textbf{Staging assessment}}\\
\midrule
OIMHS         & Macular hole                 & 4-stage   & 19   \\
ICDR          & Diabetic retinopathy                 & 5-stage   & 78   \\
SDRG          & Diabetic retinopathy                 & 5-stage   & 133  \\
\addlinespace
\multicolumn{4}{c}{\textbf{Demographics prediction}}\\
\midrule
BRSET           & Sex                 & 2-class   & 6206 \\
BRSET           & Age                 & 4-class   & 3113 \\
\bottomrule
\end{tabular}
\end{table}
\begin{table}[ht]
\centering
\captionsetup{font=large}
\caption{Balanced evaluation subset: per-task subsample sizes. Counts are per class/stage/source.}
\label{tab:balanced_subset_small}
\begin{tabular}{l l l c}
\toprule
\textbf{Parent dataset} & \textbf{Subset} & \textbf{Label schema} & \textbf{Count (per class)} \\
\midrule
\multicolumn{4}{c}{\textbf{Disease diagnosis}}\\
\midrule
Eye Disease & Multi-class              & 4-class     & 100 \\
Eye Disease & Cataract           & 2-class     & 100 \\
Eye Disease & Diabetic retinopathy & 2-class   & 100 \\
Eye Disease & Glaucoma           & 2-class     & 100 \\
\addlinespace
\multicolumn{4}{c}{\textbf{Staging assessment}}\\
\midrule
OIMHS       & Macular hole                & 4-stage     & 19  \\
\addlinespace
\bottomrule
\end{tabular}
\end{table}\newpage

\section{Details of Fine-tuning MLLM}

\begin{figure*}[ht]
\centering
\begin{minipage}{0.9\textwidth}
\begin{lstlisting}[frame=single, basicstyle=\ttfamily\footnotesize]
deepspeed llava/train/train_mem.py \
    --model_name_or_path llava-med-7b-delta \
    --data_path <DATA_PATH> \
    --vision_tower openai/clip-vit-large-patch14 \
    --mm_vision_select_layer -2 \
    --mm_use_im_start_end True \
    --bf16 True \
    --output_dir <OUTPUT_PATH> \
    --num_train_epochs 3 \
    --per_device_train_batch_size 1 \
    --per_device_eval_batch_size 4 \
    --gradient_accumulation_steps 8 \
    --evaluation_strategy "no" \
    --save_strategy "steps" \
    --save_steps 5000 \
    --save_total_limit 3 \
    --learning_rate 2e-5 \
    --weight_decay 0. \
    --warmup_ratio 0.03 \
    --lr_scheduler_type "cosine" \
    --logging_steps 1 \
    --tf32 True \
    --fsdp "full_shard auto_wrap" \
    --fsdp_transformer_layer_cls_to_wrap 'LlamaDecoderLayer' \
    --model_max_length 2048 \
    --gradient_checkpointing True \
    --lazy_preprocess True \
    --report_to wandb
\end{lstlisting}
\end{minipage}
\caption{The command to train LLaVA-Med. }
\label{fig:train_command}
\Description{Code listing showing the DeepSpeed command-line invocation and hyperparameters used to fine-tune LLaVA-Med, including model path, learning rate, batch size, and training configuration.}
\end{figure*}

\section{Demographic Subgroups Analysis}
To assess the fairness and generalizability of MLLMs across different demographic groups, we conducted a comprehensive subgroup analysis on the BRSET dataset. This analysis aims to identify potential performance disparities that may affect clinical applicability and equity in real-world deployment. 

\subsection{Dataset Selection}
The BRSET dataset was selected for subgroup analysis for several reasons.
First, BRSET provides rich and well-structured demographic annotations at the patient and exam levels, such as age and gender. These metadata allow us to precisely define and extract demographic subgroups for downstream analysis.
Second, BRSET offers a sufficient sample size across different demographic
strata. In this work, we focus on two disease diagnosis tasks with the most
balanced samples, providing adequate statistical power for subgroup comparisons:
(1) drusens detection, which identifies the presence of large drusen as an early indicator of age-related macular degeneration (AMD); and
(2) increased cup-to-disc ratio detection, which identifies abnormal
cup-to-disc ratios as a key sign of glaucoma.
Third, the BRSET dataset is highly clinically relevant. Both target conditions exhibit well-documented age- and sex-related differences in prevalence, and BRSET contains sufficient numbers of patients across these strata. This makes the dataset particularly appropriate for examining how model performance varies across demographic subgroups and for assessing potential demographic biases.
 
\subsection{Subgroup Definition and Data Preprocessing}
To enable robust and clinically meaningful subgroup comparisons, we implemented a systematic data preprocessing pipeline.\newline
\textbf{Age stratification.}
We focused on the two age groups in the BRSET dataset with the largest sample sizes:
(1) 40--60 years, representing patients in earlier screening phases where subtle abnormalities may be more challenging to detect; and
(2) 60+ years, in which disease prevalence is higher and pathological features are typically more pronounced.\newline
\textbf{Sex stratification.}
Patients were stratified into two groups based on biological sex (male vs.\ female) as recorded in the clinical metadata. Sex-based analyses are important because several retinal and optic nerve diseases show sex-related differences in prevalence and outcomes, and ensuring equitable AI performance across sexes is a critical fairness consideration.
To ensure fair and unbiased performance comparison across subgroups and disease classes, we adopted a balanced evaluation protocol: for each subgroup and disease class, we randomly sampled an equal number of positive and negative cases (1:1 ratio) using a fixed random seed. All reported metrics and statistical tests are based on these class-balanced samples, thereby controlling for potential confounding due to class imbalance.

\subsection{Evaluation Metrics and Statistical Methods}
To assess model performance across demographic subgroups, we
evaluated each model--task--subgroup combination using accuracy and F1-score.
Uncertainty in performance estimates was quantified using 95\% confidence
intervals (CIs) computed via non-parametric bootstrap. Specifically, we drew 1{,}000 bootstrap resamples with replacement, recomputed the metrics for each resample, and took the 2.5th and 97.5th percentiles of the resulting distribution as the CI bounds. 
In addition, to assess the statistical significance of performance differences between subgroups (e.g., 40--60 vs.\ 60+ years, male vs.\ female), we used chi-square tests on the binary prediction outcomes. Exact p-values are reported.

\subsection{Results and Key Findings}

 \textbf{Subgroup Performance Overview.} Table \ref{tab: age group analysis} and Table \ref{tab: gender group analysis} summarize the performance of various MLLMs across age and gender subgroups on two disease diagnosis tasks. For each model, we report accuracy and F1-score with 95\% confidence intervals (CIs) for each subgroup, as well as the p-value from a chi-square test assessing the statistical significance of performance differences between subgroups. \newline
\textbf{Age Subgroup Analysis.} The age subgroup results are shown in Table \ref{tab: age group analysis}. Across most models and both tasks, the accuracy and F1-score are highly similar between the 40--60 and 60+ age groups, with overlapping confidence intervals and p-values far above the 0.05 threshold, indicating no statistically significant performance disparity. For example, models such as InternVL-1.5-4B, InternVL-2.0-4B, and LLaVA-1.5-7B all achieve identical or near-identical metrics across age groups (e.g., accuracy $\approx$ 0.50, F1 $\approx$ 0.67, p-value = 1.0000). 
The only exception is Qwen-7B, which yields p-values below 0.05 for
both the Drusens (p = 0.0198) and Increased cup disc
(p = 0.0466) tasks. However, the corresponding effect sizes are modest:
in the Drusens task, accuracy differs by about 6\%
and F1-score by about 9\% between the 40--60 and 60+ groups.
Thus, while these differences reach statistical significance, they appear to be
of limited practical importance. Overall, we do not observe consistent evidence
of age-related performance bias across the evaluated MLLMs.\newline
\textbf{Gender Subgroup Analysis.} Similarly, the results in Table~\ref{tab: gender group analysis} show that model performance is generally consistent between male and female subgroups. For most models, accuracy and F1-score are nearly identical across sexes, and the corresponding p-values are well above 0.05, indicating no statistically significant differences. For instance, InternVL-1.5-4B, LLaVA-1.6-7B-Mistral, and Qwen-VL-Chat all report virtually
identical accuracies and F1-scores for male and female groups (p-values = 1.0000). A few models, such as InternVL-2.5-8B and Qwen-7B, exhibit small
numerical differences, but the associated p-values (all $\geq 0.24$) suggest that these fluctuations are compatible with random variation rather than systematic gender-related performance bias.\newline
\textbf{Key findings.}
These results suggest that, within the BRSET dataset and our
balanced evaluation protocol, the evaluated MLLMs do not exhibit substantial or
systematic performance disparities across age or gender subgroups on the tested disease diagnosis tasks. For the vast majority of models, the observed
differences in accuracy and F1-score between subgroups are small and not
statistically significant; even in cases with nominally significant p-values,
the corresponding effect sizes are modest. Overall, we do not find clear
evidence of strong demographic bias in model predictions under the current
setting. However, because most models still exhibit limited absolute performance, conclusions about fairness should be interpreted with caution.

\begin{table}[!ht]
    \centering
    \caption{Performance Comparison of MLLMs across age groups on disease diagnosis tasks}
    \label{tab: age group analysis}
    \resizebox{\textwidth}{!}{
    \begin{tabular}{llllll}
        \hline
        \textbf{Model} & \textbf{40-60 Accuracy (95\% CI)} & \textbf{40-60 F1 (95\% CI)} & \textbf{60+ Accuracy (95\% CI)} & \textbf{60+ F1 (95\% CI)} & \textbf{p-value} \\ \hline
        \multicolumn{6}{c}{\textbf{Drusens}}\\ \hline
        InternVL\--1.5\--2B & 0.500 [0.414, 0.579] & 0.000 [0.000, 0.000] & 0.500 [0.436, 0.569] & 0.000 [0.000, 0.000] & 1.0000 \\ 
        InternVL\--1.5\--4B & 0.500 [0.421, 0.579] & 0.667 [0.593, 0.733] & 0.500 [0.426, 0.569] & 0.667 [0.597, 0.730] & 1.0000 \\ 
        InternVL\--2.0\--2B & 0.500 [0.421, 0.579] & 0.000 [0.000, 0.000] & 0.500 [0.431, 0.569] & 0.000 [0.000, 0.000] & 1.0000 \\ 
        InternVL\--2.0\--4B & 0.500 [0.421, 0.579] & 0.667 [0.593, 0.733] & 0.500 [0.425, 0.569] & 0.667 [0.607, 0.730] & 1.0000 \\ 
        InternVL\--2.0\--8B & 0.507 [0.428, 0.586] & 0.670 [0.599, 0.736] & 0.521 [0.452, 0.591] & 0.674 [0.605, 0.735] & 0.8725 \\ 
        InternVL\--2.5\--2B & 0.513 [0.434, 0.592] & 0.673 [0.599, 0.737] & 0.489 [0.420, 0.559] & 0.647 [0.571, 0.711] & 0.7435 \\ 
        InternVL\--2.5\--4B & 0.500 [0.421, 0.579] & 0.667 [0.592, 0.733] & 0.500 [0.426, 0.569] & 0.667 [0.602, 0.725] & 1.0000 \\ 
        InternVL\--2.5\--8B & 0.592 [0.520, 0.671] & 0.693 [0.615, 0.760] & 0.585 [0.516, 0.654] & 0.705 [0.638, 0.766] & 0.9844 \\ 
        InternVL\--2.5\--MPO\--2B & 0.500 [0.414, 0.579] & 0.667 [0.593, 0.733] & 0.479 [0.410, 0.548] & 0.647 [0.581, 0.708] & 0.7786 \\ 
        InternVL\--2.5\--MPO\--4B & 0.500 [0.421, 0.579] & 0.667 [0.599, 0.733] & 0.500 [0.426, 0.569] & 0.667 [0.602, 0.725] & 1.0000 \\ 
        InternVL\--2.5\--MPO\--8B & 0.612 [0.533, 0.691] & 0.520 [0.407, 0.620] & 0.665 [0.601, 0.729] & 0.670 [0.589, 0.743] & 0.3680 \\ 
        LLaVA-1.5-7B & 0.500 [0.427, 0.579] & 0.667 [0.586, 0.733] & 0.500 [0.431, 0.574] & 0.667 [0.597, 0.725] & 1.0000 \\ 
        LLaVA-1.6-7B-Mistral & 0.500 [0.421, 0.579] & 0.667 [0.586, 0.733] & 0.500 [0.420, 0.569] & 0.667 [0.602, 0.725] & 1.0000 \\ 
        LLaVA-1.6-7B-Vicuna & 0.500 [0.421, 0.579] & 0.667 [0.593, 0.733] & 0.500 [0.431, 0.569] & 0.667 [0.602, 0.721] & 1.0000 \\
        LLaVA-1.6-13B-Vicuna & 0.500 [0.421, 0.579] & 0.667 [0.593, 0.733] & 0.500 [0.431, 0.569] & 0.667 [0.592, 0.725] & 1.0000 \\ 
        Qwen-VL-Chat & 0.500 [0.428, 0.579] & 0.667 [0.593, 0.728] & 0.500 [0.431, 0.569] & 0.667 [0.607, 0.730] & 1.0000 \\ 
        Qwen\--3B & 0.503 [0.466, 0.538] & 0.668 [0.634, 0.697] & 0.501 [0.465, 0.535] & 0.667 [0.636, 0.698] & 1.0000 \\ 
        Qwen\--7B & 0.558 [0.523, 0.591] & 0.528 [0.483, 0.570] & 0.618 [0.583, 0.653] & 0.622 [0.581, 0.662] & 0.0198 \\ 
        DeepSeek\--tiny & 0.500 [0.421, 0.586] & 0.667 [0.599, 0.728] & 0.500 [0.426, 0.569] & 0.667 [0.602, 0.725] & 1.0000 \\ 
        DeepSeek\--small & 0.500 [0.386, 0.614] & 0.667 [0.557, 0.761] & 0.500 [0.389, 0.611] & 0.667 [0.560, 0.759] & 1.0000 \\ 
        LLaVA-Med & 0.553 [0.467, 0.625] & 0.657 [0.573, 0.733] & 0.505 [0.436, 0.580] & 0.638 [0.567, 0.708] & 0.4476 \\ 
        Yi-VL & 0.490 [0.398, 0.582] & 0.653 [0.556, 0.736] & 0.467 [0.370, 0.576] & 0.632 [0.528, 0.714] & 0.8700 \\ \hline
        \multicolumn{6}{c}{\textbf{Increased cup disc}}\\ \hline
        InternVL\--1.5\--2B & 0.500 [0.445, 0.555] & 0.667 [0.618, 0.709] & 0.500 [0.442, 0.555] & 0.667 [0.616, 0.709] & 1.0000 \\ 
        InternVL\--1.5\--4B & 0.500 [0.448, 0.549] & 0.667 [0.616, 0.716] & 0.500 [0.445, 0.552] & 0.667 [0.610, 0.714] & 1.0000 \\ 
        InternVL\--2.0\--2B & 0.500 [0.448, 0.552] & 0.023 [0.000, 0.057] & 0.515 [0.460, 0.570] & 0.091 [0.036, 0.152] & 0.7505 \\ 
        InternVL\--2.0\--4B & 0.500 [0.448, 0.549] & 0.667 [0.618, 0.712] & 0.500 [0.448, 0.555] & 0.667 [0.616, 0.711] & 1.0000 \\ 
        InternVL\--2.0\--8B & 0.500 [0.448, 0.552] & 0.667 [0.618, 0.714] & 0.500 [0.445, 0.555] & 0.667 [0.619, 0.714] & 1.0000 \\ 
        InternVL\--2.5\--2B & 0.500 [0.451, 0.552] & 0.667 [0.621, 0.712] & 0.500 [0.445, 0.555] & 0.667 [0.616, 0.711] & 1.0000 \\ 
        InternVL\--2.5\--4B & 0.515 [0.462, 0.567] & 0.669 [0.618, 0.716] & 0.497 [0.442, 0.552] & 0.663 [0.611, 0.708] & 0.7050 \\ 
        InternVL\--2.5\--8B & 0.500 [0.442, 0.549] & 0.667 [0.624, 0.712] & 0.503 [0.448, 0.558] & 0.668 [0.619, 0.715] & 0.9985 \\ 
        InternVL\--2.5\--MPO\--2B & 0.500 [0.451, 0.549] & 0.667 [0.621, 0.709] & 0.500 [0.445, 0.552] & 0.667 [0.616, 0.714] & 1.0000 \\ 
        InternVL\--2.5\--MPO\--4B & 0.503 [0.448, 0.552] & 0.668 [0.615, 0.712] & 0.506 [0.451, 0.564] & 0.669 [0.623, 0.714] & 0.9956 \\ 
        InternVL\--2.5\--MPO\--8B & 0.500 [0.448, 0.552] & 0.667 [0.618, 0.712] & 0.527 [0.473, 0.579] & 0.678 [0.627, 0.723] & 0.5260 \\ 
        LLaVA-1.5-7B & 0.500 [0.448, 0.552] & 0.667 [0.616, 0.707] & 0.500 [0.448, 0.555] & 0.667 [0.616, 0.714] & 1.0000 \\ 
        LLaVA-1.6-7B-Mistral & 0.500 [0.448, 0.549] & 0.667 [0.618, 0.714] & 0.500 [0.445, 0.555] & 0.667 [0.616, 0.711] & 1.0000 \\ 
        LLaVA-1.6-7B-Vicuna & 0.500 [0.448, 0.549] & 0.667 [0.618, 0.712] & 0.500 [0.448, 0.552] & 0.667 [0.619, 0.711] & 1.0000 \\ 
        LLaVA-1.6-13B-Vicuna & 0.500 [0.448, 0.552] & 0.667 [0.613, 0.712] & 0.500 [0.442, 0.555] & 0.667 [0.619, 0.716] & 1.0000 \\ 
        Qwen-VL-Chat & 0.500 [0.448, 0.549] & 0.667 [0.616, 0.712] & 0.500 [0.445, 0.552] & 0.667 [0.613, 0.714] & 1.0000 \\ 
        Qwen\--3B & 0.500 [0.474, 0.526] & 0.667 [0.642, 0.691] & 0.500 [0.474, 0.529] & 0.667 [0.641, 0.692] & 1.0000 \\ 
        Qwen\--7B & 0.630 [0.605, 0.657] & 0.576 [0.541, 0.612] & 0.592 [0.563, 0.618] & 0.475 [0.438, 0.511] & 0.0466 \\ 
        DeepSeek\--tiny & 0.500 [0.445, 0.549] & 0.667 [0.618, 0.714] & 0.500 [0.445, 0.552] & 0.667 [0.616, 0.719] & 1.0000 \\ 
        DeepSeek\--small & 0.500 [0.448, 0.552] & 0.667 [0.618, 0.714] & 0.500 [0.445, 0.552] & 0.667 [0.619, 0.714] & 1.0000 \\ 
        LLaVA-Med & 0.500 [0.444, 0.556] & 0.667 [0.615, 0.715] & 0.500 [0.433, 0.563] & 0.667 [0.608, 0.720] & 1.0000 \\ 
        Yi-VL & 0.500 [0.443, 0.553] & 0.667 [0.614, 0.715] & 0.500 [0.444, 0.556] & 0.667 [0.618, 0.714] & 1.0000 \\ \hline
    \end{tabular}
    }
\end{table}\begin{table}[!ht]
    \centering
    \caption{Performance Comparison of MLLMs across gender groups on disease diagnosis tasks}
    \label{tab: gender group analysis}
    \resizebox{\textwidth}{!}{
    \begin{tabular}{llllll}
        \hline
        \textbf{Model} & \textbf{Male Accuracy (95\% CI)} & \textbf{Male F1 (95\% CI)} & \textbf{Female Accuracy (95\% CI)} & \textbf{Female F1 (95\% CI)} & \textbf{p-value} \\ \hline
        \multicolumn{6}{c}{\textbf{Drusens}}\\ \hline
        InternVL\--1.5\--2B & 0.500 [0.452, 0.550] & 0.000 [0.000, 0.000] & 0.500 [0.452, 0.546] & 0.000 [0.000, 0.000] & 1.0000 \\ 
        InternVL\--1.5\--4B & 0.500 [0.450, 0.546] & 0.667 [0.618, 0.708] & 0.500 [0.454, 0.548] & 0.667 [0.622, 0.708] & 1.0000 \\ 
        InternVL\--2.0\--2B & 0.500 [0.452, 0.548] & 0.000 [0.000, 0.000] & 0.500 [0.452, 0.548] & 0.000 [0.000, 0.000] & 1.0000 \\ 
        InternVL\--2.0\--4B & 0.500 [0.452, 0.548] & 0.667 [0.625, 0.708] & 0.500 [0.454, 0.553] & 0.667 [0.622, 0.708] & 1.0000 \\ 
        InternVL\--2.0\--8B & 0.505 [0.457, 0.550] & 0.668 [0.626, 0.709] & 0.512 [0.466, 0.560] & 0.672 [0.628, 0.713] & 0.8886 \\ 
        InternVL\--2.5\--2B & 0.488 [0.442, 0.538] & 0.651 [0.605, 0.695] & 0.510 [0.461, 0.553] & 0.667 [0.624, 0.710] & 0.5788 \\ 
        InternVL\--2.5\--4B & 0.500 [0.452, 0.546] & 0.667 [0.625, 0.706] & 0.500 [0.449, 0.548] & 0.667 [0.625, 0.706] & 1.0000 \\ 
        InternVL\--2.5\--8B & 0.577 [0.529, 0.625] & 0.697 [0.653, 0.737] & 0.560 [0.514, 0.609] & 0.680 [0.635, 0.721] & 0.6812 \\ 
        InternVL\--2.5\--MPO\--2B & 0.507 [0.461, 0.555] & 0.666 [0.623, 0.709] & 0.502 [0.457, 0.551] & 0.666 [0.626, 0.706] & 0.9452 \\ 
        InternVL\--2.5\--MPO\--4B & 0.500 [0.454, 0.546] & 0.667 [0.623, 0.706] & 0.500 [0.452, 0.546] & 0.667 [0.622, 0.708] & 1.0000 \\ 
        InternVL\--2.5\--MPO\--8B & 0.651 [0.603, 0.700] & 0.625 [0.570, 0.681] & 0.686 [0.643, 0.729] & 0.639 [0.576, 0.695] & 0.3253 \\ 
        LLaVA-1.5-7B & 0.500 [0.457, 0.546] & 0.667 [0.622, 0.706] & 0.500 [0.449, 0.548] & 0.667 [0.627, 0.712] & 1.0000 \\ 
        LLaVA-1.6-7B-Mistral & 0.500 [0.449, 0.548] & 0.667 [0.622, 0.706] & 0.500 [0.452, 0.546] & 0.667 [0.627, 0.708] & 1.0000 \\ 
        LLaVA-1.6-7B-Vicuna & 0.500 [0.452, 0.548] & 0.667 [0.623, 0.706] & 0.500 [0.454, 0.548] & 0.667 [0.622, 0.708] & 1.0000 \\ 
        LLaVA-1.6-13B-Vicuna & 0.500 [0.450, 0.550] & 0.667 [0.627, 0.708] & 0.500 [0.452, 0.551] & 0.667 [0.622, 0.712] & 1.0000 \\ 
        Qwen-VL-Chat & 0.500 [0.452, 0.550] & 0.667 [0.625, 0.710] & 0.500 [0.454, 0.546] & 0.667 [0.622, 0.708] & 1.0000 \\ 
        Qwen\--3B & 0.504 [0.483, 0.525] & 0.668 [0.648, 0.687] & 0.502 [0.479, 0.523] & 0.667 [0.648, 0.686] & 0.9242 \\ 
        Qwen\--7B & 0.593 [0.570, 0.615] & 0.587 [0.560, 0.611] & 0.612 [0.589, 0.633] & 0.607 [0.582, 0.632] & 0.2434 \\ 
        DeepSeek\--tiny & 0.500 [0.450, 0.548] & 0.667 [0.622, 0.708] & 0.500 [0.452, 0.551] & 0.667 [0.622, 0.706] & 1.0000 \\ 
        DeepSeek\--small & 0.500 [0.432, 0.568] & 0.667 [0.604, 0.724] & 0.500 [0.436, 0.569] & 0.667 [0.607, 0.725] & 1.0000 \\ 
        LLaVA-Med & 0.495 [0.447, 0.546] & 0.601 [0.552, 0.649] & 0.507 [0.459, 0.556] & 0.595 [0.543, 0.645] & 0.7811 \\ 
        Yi-VL & 0.466 [0.399, 0.538] & 0.629 [0.562, 0.688] & 0.505 [0.442, 0.572] & 0.660 [0.598, 0.728] & 0.4923 \\ \hline
        \multicolumn{6}{c}{\textbf{Increased cup disc}}\\ \hline
        InternVL\--1.5\--2B & 0.500 [0.461, 0.536] & 0.667 [0.632, 0.699] & 0.500 [0.462, 0.537] & 0.667 [0.633, 0.699] & 1.0000 \\ 
        InternVL\--1.5\--4B & 0.500 [0.465, 0.538] & 0.667 [0.635, 0.698] & 0.500 [0.462, 0.537] & 0.667 [0.633, 0.697] & 1.0000 \\ 
        InternVL\--2.0\--2B & 0.491 [0.455, 0.529] & 0.023 [0.000, 0.047] & 0.492 [0.456, 0.531] & 0.016 [0.000, 0.038] & 1.0000 \\ 
        InternVL\--2.0\--4B & 0.500 [0.461, 0.539] & 0.667 [0.632, 0.699] & 0.500 [0.462, 0.535] & 0.667 [0.635, 0.697] & 1.0000 \\ 
        InternVL\--2.0\--8B & 0.500 [0.461, 0.539] & 0.667 [0.631, 0.699] & 0.500 [0.463, 0.540] & 0.667 [0.633, 0.699] & 1.0000 \\ 
        InternVL\--2.5\--2B & 0.500 [0.459, 0.536] & 0.667 [0.635, 0.698] & 0.500 [0.463, 0.538] & 0.667 [0.634, 0.697] & 1.0000 \\ 
        InternVL\--2.5\--4B & 0.495 [0.458, 0.532] & 0.662 [0.628, 0.695] & 0.501 [0.463, 0.537] & 0.663 [0.628, 0.694] & 0.8683 \\ 
        InternVL\--2.5\--8B & 0.503 [0.467, 0.539] & 0.668 [0.631, 0.700] & 0.501 [0.465, 0.541] & 0.667 [0.632, 0.699] & 0.9957 \\ 
        InternVL\--2.5\--MPO\--2B & 0.500 [0.461, 0.538] & 0.667 [0.631, 0.698] & 0.500 [0.462, 0.537] & 0.667 [0.633, 0.699] & 1.0000 \\ 
        InternVL\--2.5\--MPO\--4B & 0.505 [0.468, 0.544] & 0.668 [0.634, 0.700] & 0.501 [0.463, 0.540] & 0.667 [0.632, 0.695] & 0.9511 \\ 
        InternVL\--2.5\--MPO\--8B & 0.524 [0.489, 0.562] & 0.678 [0.645, 0.712] & 0.510 [0.472, 0.545] & 0.670 [0.639, 0.701] & 0.6351 \\ 
        LLaVA-1.5-7B & 0.500 [0.464, 0.539] & 0.667 [0.632, 0.698] & 0.500 [0.460, 0.537] & 0.667 [0.633, 0.699] & 1.0000 \\ 
        LLaVA-1.6-7B-Mistral & 0.500 [0.461, 0.538] & 0.667 [0.631, 0.699] & 0.500 [0.460, 0.537] & 0.667 [0.632, 0.695] & 1.0000 \\ 
        LLaVA-1.6-7B-Vicuna & 0.500 [0.461, 0.539] & 0.667 [0.632, 0.699] & 0.500 [0.462, 0.534] & 0.667 [0.635, 0.697] & 1.0000 \\ 
        LLaVA-1.6-13B-Vicuna & 0.500 [0.461, 0.538] & 0.667 [0.631, 0.699] & 0.500 [0.462, 0.535] & 0.667 [0.632, 0.700] & 1.0000 \\ 
        Qwen-VL-Chat & 0.500 [0.459, 0.535] & 0.667 [0.634, 0.699] & 0.500 [0.463, 0.538] & 0.667 [0.632, 0.697] & 1.0000 \\ 
        Qwen\--3B & 0.500 [0.482, 0.518] & 0.667 [0.650, 0.682] & 0.500 [0.481, 0.519] & 0.667 [0.650, 0.683] & 1.0000 \\ 
        Qwen\--7B & 0.592 [0.573, 0.611] & 0.546 [0.522, 0.570] & 0.581 [0.562, 0.601] & 0.517 [0.492, 0.541] & 0.4597 \\ 
        DeepSeek\--tiny & 0.500 [0.461, 0.538] & 0.667 [0.632, 0.698] & 0.500 [0.463, 0.538] & 0.667 [0.635, 0.700] & 1.0000 \\ 
        DeepSeek\--small & 0.500 [0.465, 0.538] & 0.667 [0.631, 0.699] & 0.500 [0.462, 0.537] & 0.667 [0.632, 0.699] & 1.0000 \\ 
        LLaVA-Med & 0.500 [0.450, 0.547] & 0.667 [0.621, 0.708] & 0.500 [0.458, 0.546] & 0.667 [0.624, 0.707] & 1.0000 \\ 
        Yi-VL & 0.500 [0.459, 0.541] & 0.667 [0.629, 0.702] & 0.500 [0.462, 0.538] & 0.667 [0.633, 0.700] & 1.0000 \\ \hline
    \end{tabular}
    }
\end{table}

\end{document}